\documentclass[runningheads]{llncs}
\usepackage{graphicx}

\usepackage{tikz}
\usepackage{comment}
\usepackage{amsmath,amssymb} %
\usepackage{color}
\usepackage{multirow, multicol}
\usepackage[accsupp]{axessibility}  %

\newcommand{\rev}[1]{{\color{black}#1}}

\begin{document}
\pagestyle{headings}
\mainmatter
\def\ECCVSubNumber{5134}  %

\title{Are High-Resolution Event Cameras \\Really Needed?} %

\titlerunning{Are High-Resolution Event Cameras Really Needed?}
\author{Daniel Gehrig \and
Davide Scaramuzza}
\institute{Dept. of Informatics, Univ. of Zurich and \\
Dept. of Neuroinformatics, Univ. of Zurich and ETH Zurich\\
\email{\{dgehrig,sdavide\}@ifi.uzh.ch}}
\maketitle

\begin{abstract}
Due to their outstanding properties in challenging conditions, event cameras have become indispensable in a wide range of applications, ranging from automotive, computational photography, and SLAM.  
However, as further improvements are made to the sensor design, modern event cameras are trending toward higher and higher sensor resolutions, which result in higher bandwidth and computational requirements on downstream tasks. 
Despite this trend, the benefits of using high-resolution event cameras to solve standard computer vision tasks are still not clear. 
In this work, we report the surprising discovery that, in low-illumination conditions and at high speeds, low-resolution cameras can outperform high-resolution ones, while requiring a significantly lower bandwidth.  
We provide both empirical and theoretical evidence for this claim, which indicates that high-resolution event cameras exhibit higher per-pixel event rates, leading to higher temporal noise in low-illumination conditions and at high speeds. 
As a result, \rev{in most cases}, high-resolution event cameras show a lower task performance, compared to lower resolution sensors in these conditions.
We empirically validate our findings across several tasks, namely image reconstruction, optical flow estimation, and camera pose tracking,
both on synthetic and real data.
We believe that these findings will provide important guidelines for future trends in event camera development.

\keywords{Methodology, and Theory, Low-level Vision}
\end{abstract}

\section*{Multimedia Material}
For videos and more, visit our project page at \url{https://uzh-rpg.github.io/eres/}.

\section{Introduction}
\begin{figure}
    \centering
    \includegraphics[width=0.8\linewidth]{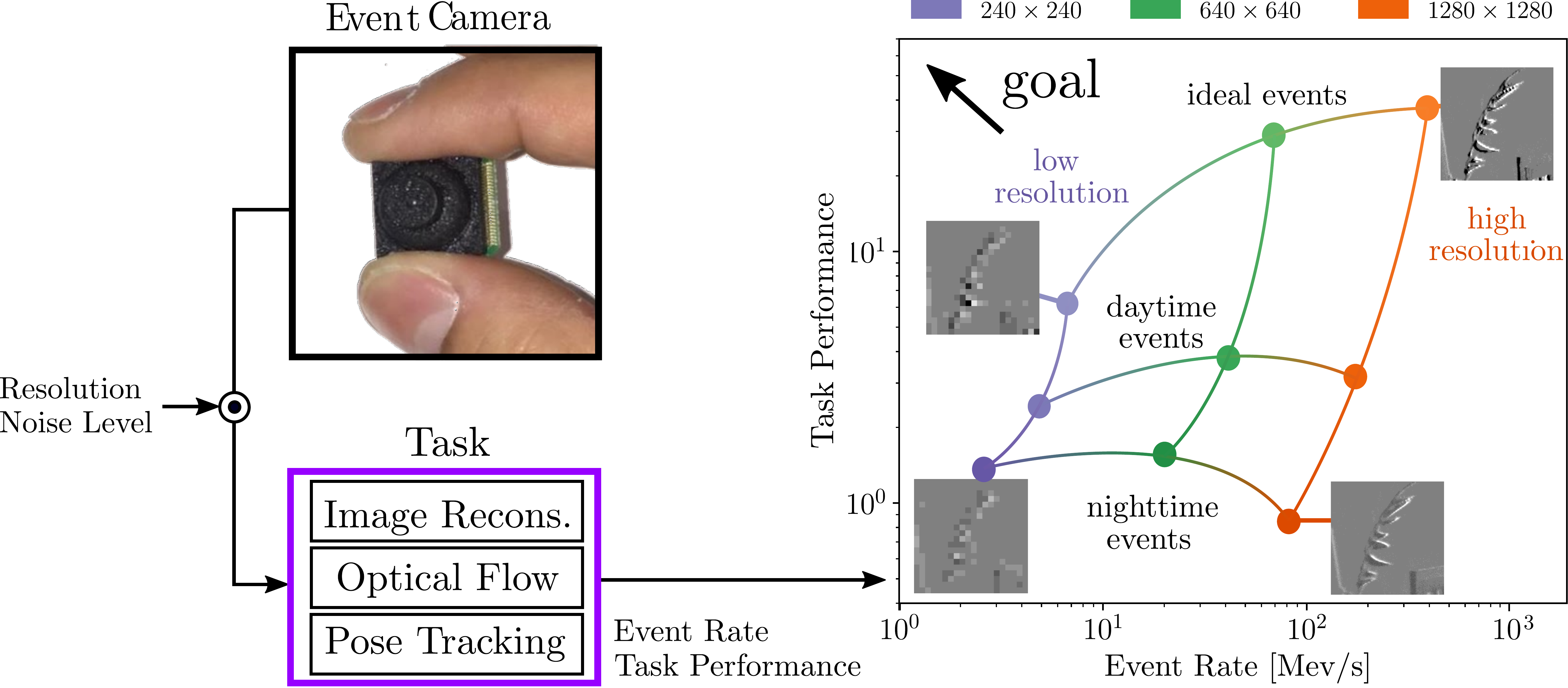}
    \caption{We study the impact of camera resolution on downstream task performance and bandwidth requirements. Intuitively, in ideal settings (top curve) using higher resolution sensors leads to higher performance, but also higher event rates, which cause higher bandwidths.  
    However, in realistic noise conditions (bottom curves) this is no longer the case. 
    We find that especially at high speeds and in low-illumination conditions, using high-resolution cameras significantly decreases the task performance. 
    In these settings, lower resolutions are more viable in terms of performance and bandwidth.}\label{fig:eyecatcher}%
    \vspace{-2ex}
\end{figure}

Event cameras are neuromorphic vision sensors that work radically different than standard frame-based cameras. 
Instead of measuring synchronous intensity frames at fixed time intervals, event cameras only measure the \emph{changes} in logarithmic intensity and do this asynchronously and with microsecond-level temporal resolution.
The resulting output is a stream of asynchronous \emph{events}, where each event encodes the pixel location, 
time and sign (polarity) of the intensity change.
Due to their working principle, event cameras have several advantages over standard cameras, including a 
higher dynamic range (140 dB vs. 60 dB), low power consumption (milliwatts instead of watts), low motion blur, and microsecond-level latency and temporal resolution. 

Due to these properties, applications and research using event cameras has spread to various fields, including robotics\cite{Glover17iros,Falanga20science,Sanket20icra,Sun21ral,Hagenaars20ral}, 
tactile sensing\cite{Baghaei20tim,Taunyazov20rss}, 
high-speed control\cite{Dimitrova20icra,Vitale21icra}, %
driving\cite{Gehrig21ral,Sironi18cvpr,Zhu18rss}, 
space \cite{Cohen19jas}
and computational photography\cite{Rebecq19pami,Tulyakov21cvpr,Bardow16cvpr,Zhang21cvpr}. 
Since their surge in popularity, event-camera technology has continuously improved, leading to several commercially available sensors, summarized in Tab.\ref{tab:event_cameras} (borrowed from the survey in \cite{Gallego20pami}).
These improvements include better noise characteristics and higher sensor resolution.
While early event cameras such as the DVS128 had a resolution 128$\times$128\cite{Lichtsteiner08ssc}, modern event cameras now go up to 1280$\times$960\cite{Suh20iscas}.

While higher sensor resolutions have enabled new applications, such as computational photography \cite{Rebecq19pami,Tulyakov21cvpr}, they are increasingly placing a burden on the end-user both in terms of data bandwidth and computational requirements of downstream systems\cite{Muglikar21threedva,Muglikar21threedvb}. %
This is because, as resolution increases, so does the resulting event rate.
Common issues that arise with modern high-resolution event cameras are slow readout rates and output bus saturation\cite{Gallego20pami}, which both lead to event timestamp perturbations and loss of events.
While readout rates have been steadily increasing from 2 MHz \cite{Lichtsteiner08ssc} to 1200 MHz \cite{Suh20iscas}, they require sophisticated readout schemes to reduce overhead. 
For this reason, some solutions use hardware-integrated filters to reduce the event rate\cite{Finateu20isscc}. %
However, while existing strategies reduce the overall bandwidth, they are not lossless, and, thus, either skip events or introduce systematic noise on the event timestamps. 
Both can have deleterious effects on the downstream task.

This raises the question, of what the true benefits of high-resolution cameras are.
In this work, we shed light on this question by empirically and theoretically studying the effect of event camera resolution on three downstream tasks, namely \emph{(i)} image reconstruction, \emph{(ii)} optical flow estimation, and \emph{(iii)} camera pose tracking.
For each task, we investigate the trade-off between task performance and event rate in ideal, daytime and nighttime conditions (Fig.\ref{fig:eyecatcher}). 
We find both on synthetic and real data that higher resolution event cameras exhibit higher \emph{per-pixel event rates}, which makes them less robust to high-speed motion and more susceptible to temporal noise induced by low illumination conditions.
This leads to the surprising discovery that, in these conditions, lower resolution event cameras exhibit better performance than their high-resolution counterparts, while also producing fewer events. We demonstrate that this conclusion generalizes across several tasks and both learning- and model-based methods.
\rev{Additionally, we find that, when including noisy nighttime data in the training set, learning-based methods can overcome this trend. Moreover, contrast-maximization-based methods~\cite{Gallego18cvpr} only show this trend at high speeds, while photometric-based methods~\cite{Gallego17pami} already degrade at low speeds.}
We believe that these findings will act as important guidelines for future trends in event camera manufacturing, especially when moving toward higher resolution sensors.
Our contributions are:
\begin{itemize}
    \item We theoretically and empirically study the effect of higher resolutions on per-pixel event rates and show that event rates with increasing sensor resolution.
    \item We study the effect of event camera resolution on three important computer vision tasks. We show that, both in daytime and nighttime conditions and during high-speed motions, tasks performed with high-resolution cameras show a significant performance drop, which is smaller for lower resolutions. 
    \item We thus identify event-camera resolutions which outperform high-resolution event cameras in terms of task performance, while requiring less bandwidth and do so over a variety of model-based and learning-based methods, as well as in simulation and on real data.  
\end{itemize}

\section{Related Work}

\begin{table}[t!]
    \centering
\resizebox{0.8\linewidth}{!}{
    \begin{tabular}{l|l|llll}
    \hline
    \textbf{Supplier}                   & \textbf{Camera}     & \textbf{Year,}     & \textbf{Resolution}  & \textbf{Interface} & \textbf{Max. Bandwidth} \\ 
                                        &                           & \textbf{Reference} & \textbf{(pixels)}   && \textbf{(Mev/s)}                    \\ \hline
    \multirow{4}{*}{iniVation} & DVS128           & 2008\cite{Lichtsteiner08ssc}     & $128\times 128$     & USB 2                      &1      \\
                               & DAVIS240         & 2014\cite{Brandli14ssc}          & $240\times 180$     & USB 2                     &12      \\
                               & DAVIS346         & 2017            & $346\times 260$                      & USB 3                     &12      \\
                               & DVXPlorer        & 2020            & $640\times 480$                      & USB 3                    &165      \\ \hline
    \multirow{4}{*}{Prophesee} & ATIS             & 2011\cite{Posch11ssc}            & $304 \times 240$     & -                      & -         \\
                               & Gen3 CD          & 2017\cite{propheseeevk}          & $640 \times 480$     & USB 3                     &66      \\
                               & Gen3 ATIS        & 2017\cite{propheseeevk}          & $480 \times 360$     & USB 3                     &66      \\
                               & Gen4 CD          & 2020\cite{Finateu20isscc}        & $1280\times 720$    & USB 3                   &1066      \\ \hline
    \multirow{3}{*}{Samsung}   & DVS-Gen2         & 2017\cite{Son17isscc}            & $640 \times 480$     & USB 2                    &300      \\
                               & DVS-Gen3         & 2018\cite{Ryu19cvprw}            & $640 \times 480$     & USB 3                    &600      \\
                               & DVS-Gen4         & 2020\cite{Suh20iscas}            & $1280\times 960$    & USB 3                   &1200      \\ \hline
    \multirow{2}{*}{CelePixel} & CeleX-IV         & 2017\cite{Guo17iscas}            & $768 \times 640$     & -                    & 200         \\
                               & CeleX-V          & 2019\cite{Chen19cvprw}           & $1280\times 800$    & -                    & 140         \\ \hline
    Insightness                & Rhino 3          & 2018\cite{Insightness19rino}     & $320 \times 262$     & USB 2                     & 50     \\ \hline
    \end{tabular}}
    \caption{Event camera comparison. Courtesy of \cite{Gallego20pami}.}\label{tab:event_cameras}
    \vspace{-4ex}
\end{table}

Since the first commercial event camera in 2008\cite{Lichtsteiner08ssc}, several new sensors appeared, (Tab. \ref{tab:event_cameras}) which drove the development of different application domains. 
While early works used the DVS\cite{Lichtsteiner08ssc}, ATIS\cite{Posch11ssc} and DAVIS240\cite{Brandli14ssc} to solve simple tasks like low degree-of-freedom SLAM or object recognition,
more modern approaches use higher resolution event cameras such as the Prophesee Gen3, Rhino 3, and Samsung Gen3 which provide more fine-grained detail in the event stream. 
Most recently, a race to increase sensor resolution culminated in the Celex-V, Prophesee Gen4, and Samsung DVS-Gen4, all featuring approximately one-megapixel resolution. 
These sensors enable applications in computational photography, where higher resolutions directly translate to more fine-grained detail. 
However, higher resolutions result in higher event rates, which place burdens on the data bus, readout systems, and downstream tasks.
These include readout congestion and bus saturation, both of which can introduce significant delays or noise on event timestamps, and even dropped events\cite{Gallego20pami}.
In the next section, we discuss strategies to overcome this challenge.\\

\textbf{Event Filtering and Compression:}
Higher event rates directly translate to higher storage requirements. 
To counteract these, lossy and lossless event compression techniques were introduced\cite{Banerjee21icip,Khan2020ieee}. 
However, these methods do not work online and are thus not able to mitigate event distortions at recording time.

Modern event cameras such as the Prophesee Gen4\cite{Finateu20isscc} include sophisticated read-out schemes such as row-wise arbitration which reduce per-event data transmission but introduce systematic errors in the event timestamps. Recently, event rate controllers were introduced which aim at reducing the event rate while recording. 
They work by randomly skipping events \cite{Finateu20isscc} or by tuning camera parameters during recording\cite{Delbruck21cvprw}.   
While these controllers can alleviate bandwidth requirements, they do so by directly affecting the events.
Thus algorithms and event rate controllers need to be co-designed, thereby increasing complexity which inhibits scaling to more complex tasks.
Instead of filtering the event stream we seek to understand the benefits of high-resolution cameras and their trade-offs in terms of downstream task performance and bandwidth.\\

\textbf{Power-Performance Analysis:}
The tradeoff between task performance and data bandwidth was initially studied in \cite{Censi15icra} in the context of a Power-Performance analysis. They compared frame- and event-based sensors in terms of their ability to reconstruct the continuous-time intensity profile at a single pixel (Performance), and their required bandwidth (Power). %
However, this analysis was limited to a simple simulated toy example\footnote{The studied intensity signal was modeled as a random process that did not resemble real data and did not model sensor noise.} which did not consider sensor noise. 
In this work, we generalize this analysis to event cameras at different resolutions and in different conditions and study them in terms of their task performance and event rate (Fig.\ref{fig:eyecatcher}).
We show results for various tasks and methods, and both in simulation and on real data. 

\section{Approach}
\begin{figure*}[t!]
    \centering
    \begin{tabular}{ccc}
        \includegraphics[height=0.3\textwidth]{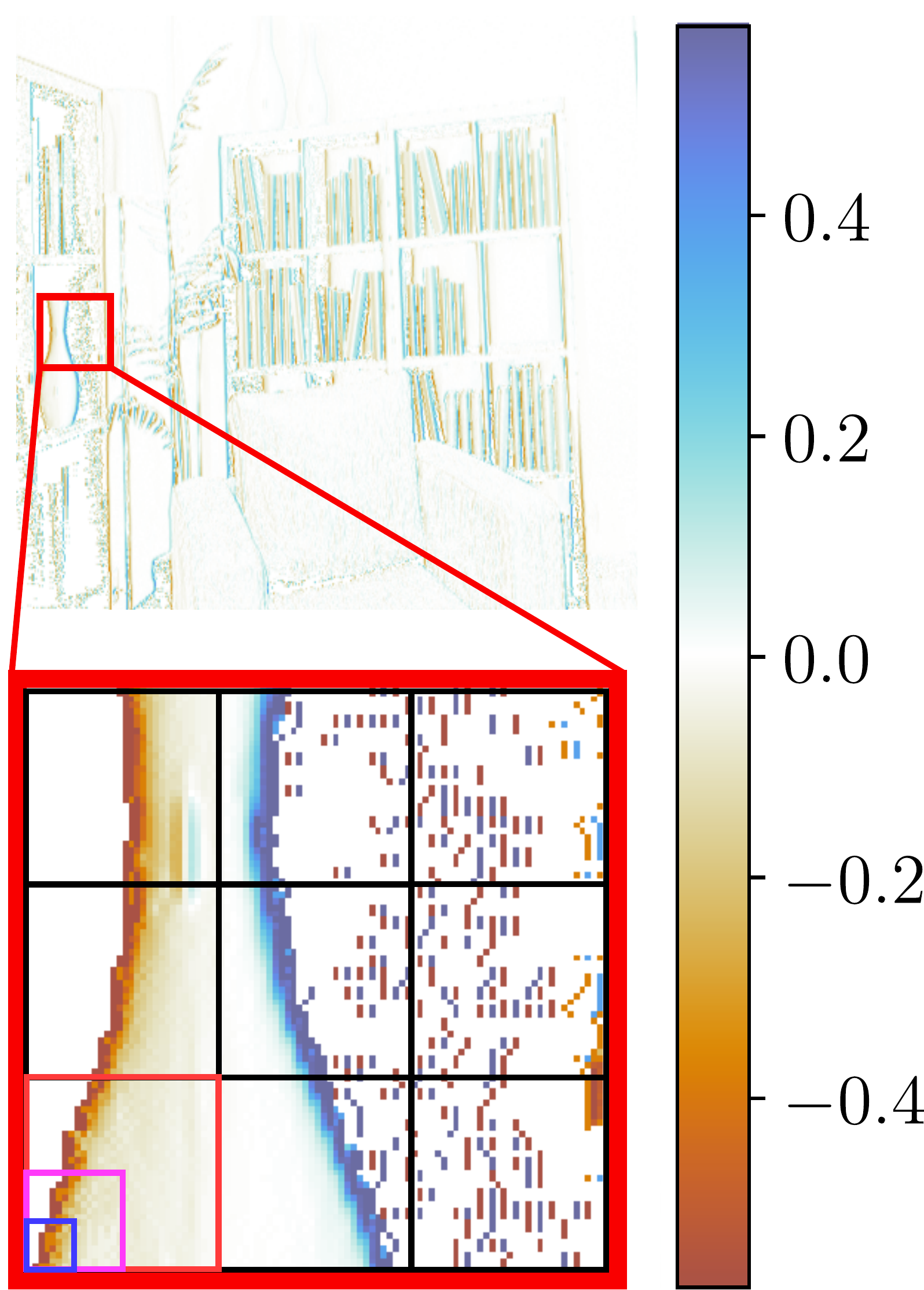}&
        \includegraphics[height=0.3\textwidth]{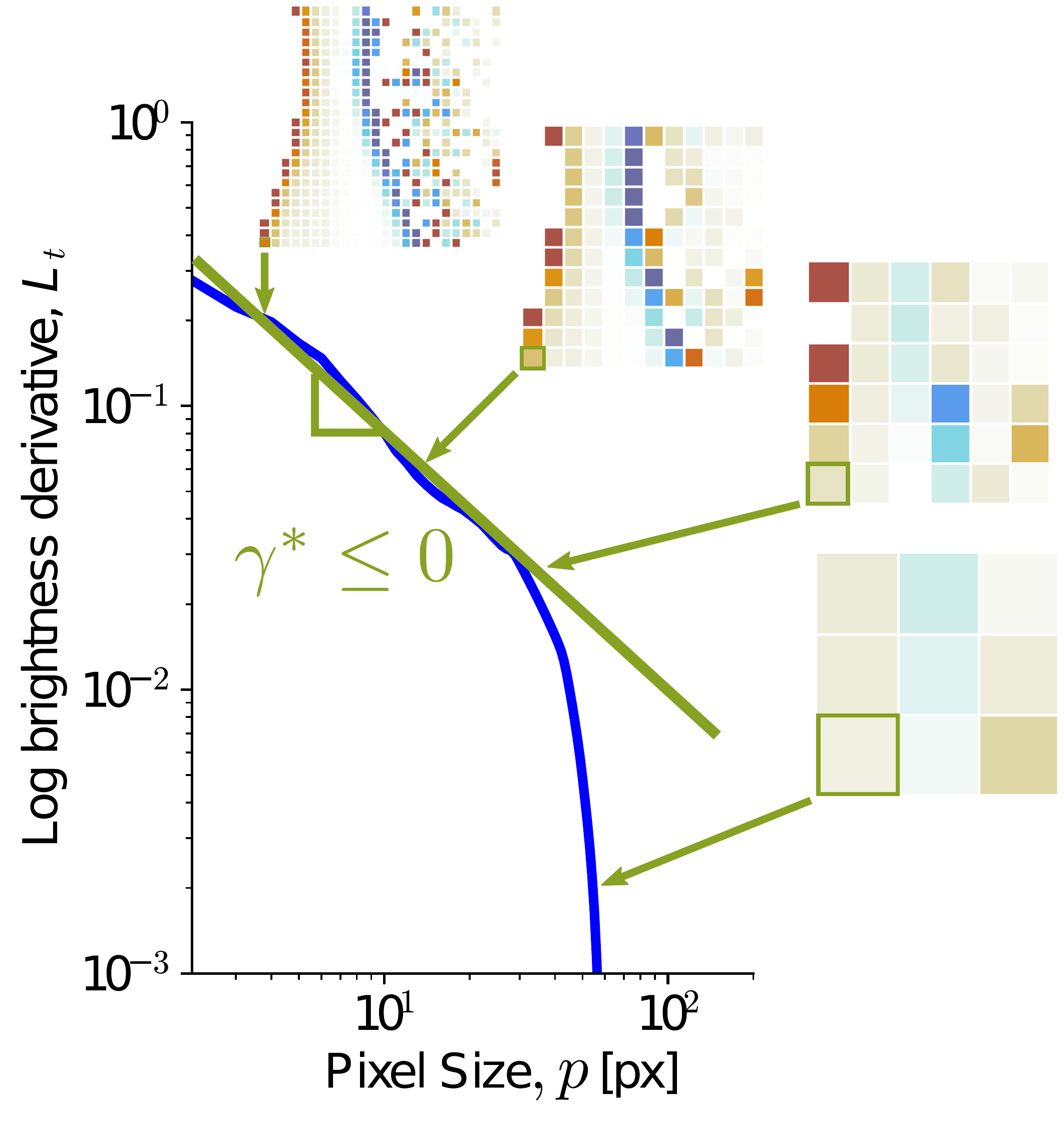}&\quad
        \includegraphics[height=0.3\textwidth]{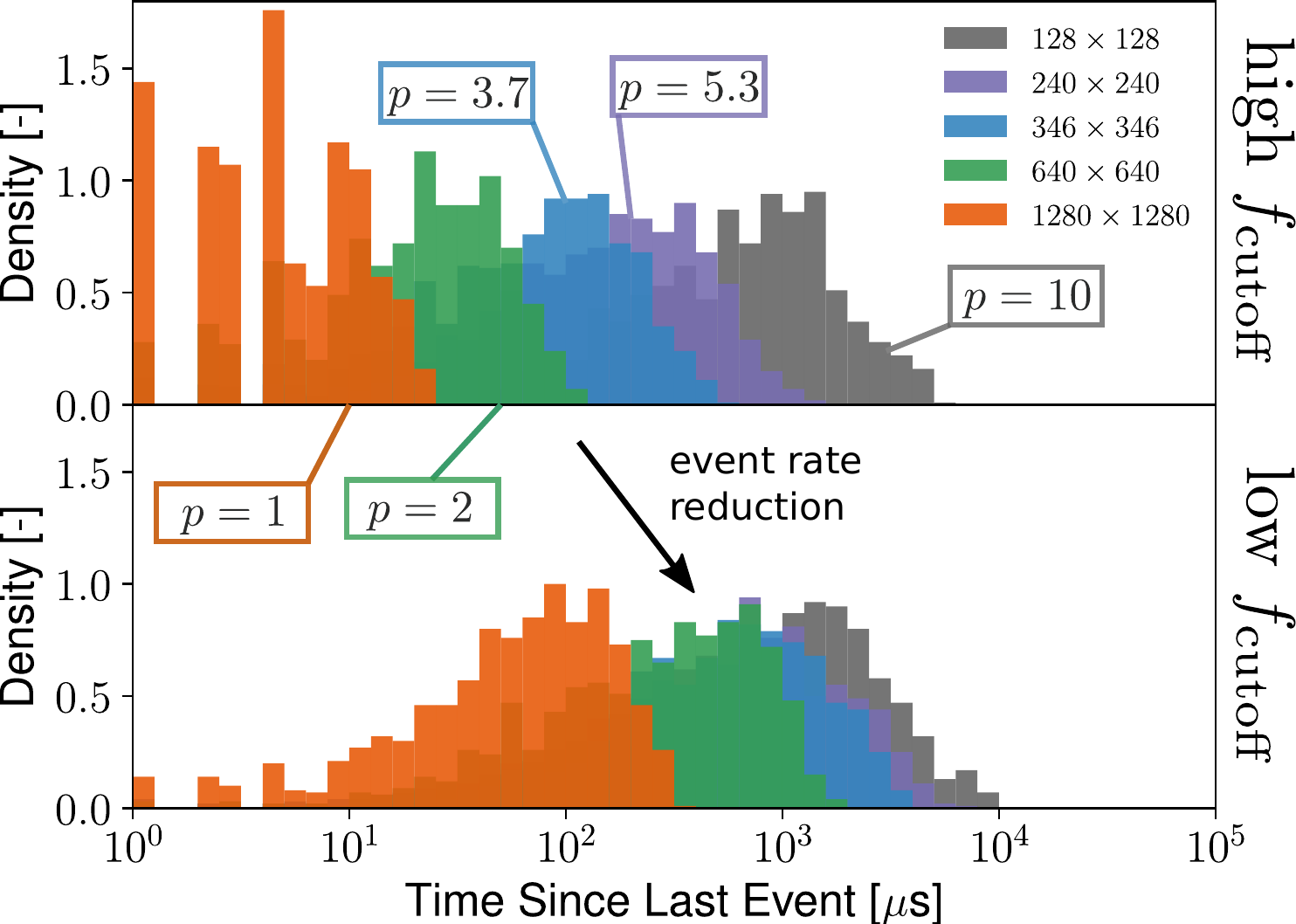}\\
        (a) derivative $L_t$ & (b) $L_t$ vs. pixel size &(c) time between consecutive events        
    \end{tabular}
    \caption{To understand the event rate at different resolutions we study the log brightness derivative $L_t$ of a given scene (a). Brightness changes are mostly confined to edges, making them semi-dense. We consider the brightness derivative centered at an edge pixel of varying size  (b), and find that as pixel size decreases, more details become visible which increase $L_t$ (b), leading to a power law with slope $\gamma^*<0$ (related to the fractal dimension\cite{Messikommer20eccv} at small pixel sizes).
    Thus, to satisfy Eq. \eqref{eq:gen_model_linear} high resolution event cameras must trigger events at higher rates (c, top), making them more sensitive to temporal effects, such as pixel cutoff frequency or high-speed motion (c, bottom). }\label{fig:scaling_law}\vspace{-2ex}
\end{figure*}
In the first step, we will review the event generation model before studying events at higher resolution.
Finally, we will introduce baselines which we will use to quantify task performance. 

\textbf{Event Generation:}
\label{sec:gen_model}
Event cameras have independent pixels $\mathbf{x}=(x,y)$ which trigger an event whenever their internal voltage $L(\mathbf{x}, t)$, caused by the illumination incident on that pixel,
exceeds a certain threshold $C$, called the contrast threshold. 
The resulting events are four tuples $e_k=(\mathbf{x}_k, t_k, p_k)$, each encoding the location $\mathbf{x}_k$, time $t_k$ and sign (polarity) $p_k\in \{-1,1\}$ of this change.
The event trigger condition can be formalized as follow 
\begin{equation}
    \label{eq:gen_model}
    p_k(L(\mathbf{x}_k,t_k) - L(\mathbf{x}_k,t-\Delta t_k)) > C,
\end{equation}  
which describes the ideal event generation model \cite{Gallego15arxiv,Gallego17pami}. 
Here $\Delta t_k$ is the time since the last event at pixel $\textbf{x}_k$.
By linearizing $L$, Eq. \eqref{eq:gen_model} can be rewritten as 
\begin{equation}
    \label{eq:gen_model_linear}
    p_k L_t(\mathbf{x}_k,t_k)\Delta t_k > C,
\end{equation}
with $L_t = \frac{\partial L}{\partial t}$. 
While in the ideal setting the internal voltage level is equated with the logarithmic brightness $L(\mathbf{x},t)\doteq \log I(\mathbf{x},t)$, 
more recently, a more realistic event-camera model has been introduced which models
this voltage as a second order system, dependent on the cutoff frequency $f_{\text{cutoff}}$~\cite{Hu21cvprw}.
It determines the rate at which event pixels track an input brightness signal, and depends on the ambient illumination conditions.
In darker settings $f_{\text{cutoff}}$ is lower, causing event pixels to become slow and generate characteristic ``motion blur patterns",
\rev{which can be viewed in Fig. 3 of \cite{Hu21cvprw}.}%

\textbf{Event Generation at Different Resolutions:}
\label{sec:sec:gen_model_multi_scale}
While Eq. \eqref{eq:gen_model_linear} assumes that per-pixel scene irradiance is measured in a single direction, it is integrated
over a steric angle, which spans a single pixel. 
As camera resolution increases, this pixel area decreases, leading to a more fine-grained perception of the environment. 
This integration is followed by a subsampling of the image. Let $E$ denote the irradiance map at a high resolution and $\textbf{x}$ denote the coordinates at this high resolution. 
Then log intensity at low resolution can be computed in two steps:
\begin{align}
    \label{eq:int_irradiance}
    L(\textbf{x},t) &= \log I(\textbf{x},t) = \log [(k*E)(\textbf{x},t)], \\
    \nonumber\text{with }\quad k(\textbf{x}) &= 
    \begin{cases}
        \frac{1}{p^2},\quad -\frac{p}{2}\leq x,y \leq\frac{p}{2}\\
        0 \quad\text{ else}
    \end{cases}
\end{align}
where the scene irradiance $E$ is convolved with a box filter $k$ with sides $p$ corresponding with the pixel size. Followed by a subsampling to a lower resolution with coordinates $\textbf{x}_l$
\begin{equation}
    L_l(\textbf{x}_l,t)=L(p \textbf{x}_l,t)
\end{equation}

We consider the event rate at a fixed pixel  $\textbf{x}$, with coordinates $\textbf{x}_l=\frac{\textbf{x}}{p}$ for pixel size $p$, by studying the temporal derivative $L_t$, at that pixel:
\begin{equation}
    \label{eq:log_deriv_bright}
    L_t(\textbf{x},t) = \frac{\partial \log I(\textbf{x},t)}{\partial t} = \frac{I_t(\textbf{x},t)}{I(\textbf{x},t)}=\frac{(k*E_t)(\textbf{x},t)}{(k*E)(\textbf{x},t)}
\end{equation}
We observe that $E_t$ is sparse, with high values close to occlusions and contrast changes (Fig. \ref{fig:scaling_law} (a), top). 
Thus aggregations of larger areas diminish, leading to lower values at higher pixel sizes.
Plotting $L_t$ centered at an individual pixel, reveals a power-law for low pixel sizes, \emph{i.e.} high resolutions (Fig. \ref{fig:scaling_law} b)  
\begin{equation}
    L_t(\textbf{x},t) \propto \frac{1}{p^{\gamma^*(\textbf{x})}}\quad \text{ as } p\rightarrow 0.
\end{equation}
Here $\gamma^*$ is related to the fractal dimensions of $E_t$ and $E$ \cite{Messikommer20eccv}.
When an event is triggered, the constraint \eqref{eq:gen_model_linear} must be satisfied, and thus we find that the time between consecutive events at a single pixel, $\Delta t_k$, must satisfy
\begin{equation}
    \label{eq:timestamp_scaling}
    \Delta t_k = \frac{p_k C} {L_t(\textbf{x}_k,t_k)} \propto p^{\gamma^*(\textbf{x}_k,t_k)}\quad \text{ as } p\rightarrow 0.
\end{equation}
This means that as sensor resolution increases, the time window between consecutive events at a specific pixel must decrease. \rev{This assumes 
a constant contrast threshold. If the contrast theshold is simultaneously increased, this effect can be counteracted but results in a loss of detail. 
However, in this work we do not study the effect of contrast threshold variations, and leave it for future work.} 
We verify this relation in Fig. \ref{fig:scaling_law} (c), where we show histograms of $\Delta t_k$ across common event camera resolutions.
We see that as for smaller pixels, \emph{i.e} higher resolutions, the histograms shift to lower $\Delta t_k$.
The mode of the distribution shifts high $\Delta t$  to low $\Delta t$s at $p$ decreases.
Due to the increased event rate, small perturbations of event timestamps, caused by slow pixel response times lead to more significant effects on following events. 
This indicates that high-resolution event cameras are more susceptible to temporal noise effects caused by cutoff frequency and high speed motions (see Fig.~\ref{fig:scaling_law} (c) bottom).

\subsection{Tasks}
\label{sec:tasks}

\textbf{Image Reconstruction}
We use simple event integration as a baseline for image reconstruction, directly leveraging Eq. \eqref{eq:gen_model}. 
Given a high resolution image $L_{h,0}(\textbf{x})$ at time $t_0$ and low resolution events $\mathcal{E}_l=\{e_i\}_{i=1}^{N}$ 
 we can produce a prediction of the image at $t_1$ as follows:
\begin{align}
    \label{eq:event_quality}
    \hat L_{1,h}(\textbf{x}) = L_{0,h}(\textbf{x})+\mathcal{U}\left(\Delta L_l \right)(\textbf{x})\\
    \nonumber\Delta L_l(\textbf{x}_l) = \sum_{i=0}^N p_i C\delta(\textbf{x}_l-\textbf{x}_i)
\end{align}    
Where $\delta$ denotes the Kronecker delta and $\mathcal{U}$ denotes the upsampling operator, which upsamples the low-resolution brightness increment 
to the same resolution as the high-resolution images. \rev{In what follows, we will fix the scale of the high-resolution image and vary the 
resolution of the events. This way, we can compare predictions across event camera scales, without changing the ground-truth image $L_{1,h}(\textbf{x})$}.
Since this method depends on $C$ which is typically unknown we will only use it in simulation.
However, due to its direct relationship to the event generation model, it serves as a good baseline for evaluating the quality of the events. In the following sections, we will refer to it as the \emph{Event Integration} (EI) baseline.\\

\textbf{Optical Flow Estimation}
Following the event generation model in Eq. \eqref{eq:gen_model} we see that consecutive events at the same pixel measure a brightness change of 
$\pm C$. 
We design a flow method that enforces this constraint. 
Let $L_l(\textbf{x})$ be a low resolution image at time $t_0$ and let us focus on low resolution events triggered in a small patch and time window $\Delta T$ after $t_0$.
We assume that the optical flow $\textbf{v}$ is approximately constant within this spatio-temporal volume. 
Therefore, locally the brightness at any time $t$
can be approximated with 
\begin{equation}\label{eq:brightness_over_time}
    L(\textbf{x}_l,t) = L_l(\textbf{x}_l-\textbf{v}t)
\end{equation}
Where for simplicity we measure $t$ relative to $t_0$. 
By inserting Eq. \eqref{eq:brightness_over_time} into the generative model in Eq. \eqref{eq:gen_model} and enforcing equality we have for a single event
\begin{equation}\label{eq:photometric_constraint_single_event}
    p_k (L_{l}(\textbf{x}_k-\textbf{v} t_k) - L_{l}(\textbf{x}_k-\textbf{v}(t_k-\Delta t_k))) = C
\end{equation}
Eq. \eqref{eq:photometric_constraint_single_event} forces the difference of the intensities
at coordinates $\textbf{x}_k-\textbf{v}t_k$ and $\textbf{x}_k-\textbf{v}(t_k - \Delta t_k)$ to be equal to $\pm C$. 
We denote $\Delta L_{l}^k(\textbf{v})\doteq L_{l}(\textbf{x}_k-\textbf{v}t_k) - L_{l}(\textbf{x}_k-\textbf{v}(t_k-\Delta t_k))$
and then stack the above equation for several events resulting in 
\begin{equation}
    \label{eq:optical_flow_objective}
    \hat{\textbf{v}} = \arg \min_\textbf{v}\Vert\Delta \textbf{L}_l(\textbf{v})\odot\textbf{p}-\textbf{1}C\Vert^2
\end{equation} 
With $\odot$ denoting element-wise multiplication. We minimize Eq. \eqref{eq:optical_flow_objective} in two steps. We first solve for the optimal contrast threshold, $\hat{C}$, with the closed form solution
\begin{align}
    \hat C &= \frac{\textbf{p}^T\Delta\textbf{L}_l(\textbf{v})}{\textbf{p}^T\textbf{p}},
\end{align}
 before resubstituting and minimizing the original objective for $\textbf{v}$ using a first-order method.
We call this method \emph{Event-based Photometric Flow} (EPF).\\

\textbf{Camera Pose Tracking}
Camera pose tracking estimates the camera pose given a photometric depth map, as in \cite{Gallego17pami}.
We develop a pose tracker, based on \cite{Gallego17pami} and \cite{Mueggler18tro}. 
We parametrize the camera trajectory using a cumulative B-spline\cite{Mueggler18tro} $T(t)$, parametrized by coordinates $\xi$ and use the events, together with the photometric depth map to constrain this spline.
Again, we use Eq. \eqref{eq:gen_model}, to set up a constraint for each event, but modify Eq. \eqref{eq:brightness_over_time} to fit into  
a pose tracking setting. 
We use a low resolution frame $L_l(\textbf{x})$ together with a depthmap $D(\textbf{x})$ at time $t_0$, and low resolution events. 
Here, $L_l(\textbf{x})$ plays the role of reference view. 
The intensity at each event is found by backprojecting the event at position $\textbf{x}_k$ and camera pose T($t_k$) and reprojecting into 
the reference view.
\begin{align}
    L(\textbf{x}, t) &= L_l(\textbf{x}')\\
    \nonumber\textbf{x}'&=\pi(\textbf{X},K, T_0)\\
    \nonumber\textbf{X}&=\pi^{-1}(\textbf{x},Z,K, T(t))
\end{align}
Where $\pi$ projects and backprojects pixel coordinates according to camera matrix $K$, pose $T(t)$, and depth $Z$. Since $T$ is a spline, it can be sampled at arbitrary times.
$T_0$ stands for the camera pose and $\textbf{X}$ denotes the 3-D point corresponding to the event.
We assume planar scenes, to avoid the need for raytracing, and thus find a closed-form solution for the depth $Z$ for each event. 
Thus for a single event 
\begin{equation}
    p_k(L_l(\textbf{x}'_1(\xi) - L_l(\textbf{x}'_0(\xi)) = C,
\end{equation}
where each point $\textbf{x}'_{0/1}$ depends on the spline $\xi$.
Stacking these residuals we get
\begin{equation}
    \label{eq:pose_tracking} 
    \xi = \arg \min_\xi \Vert \Delta \textbf{L}_l(\mathbf{\xi})\odot\textbf{p}-\textbf{1}\hat{C}\Vert^2
\end{equation}
which we solve using the non-linear least squares solver Ceres\cite{ceres-solver}. In what follows, we will refer to this method as \emph{Event-based Photometric Pose Tracker} (EPPT).

\begin{figure*}[t!]
    \centering
    \begin{tabular}{ccc}
        \includegraphics[height=3cm]{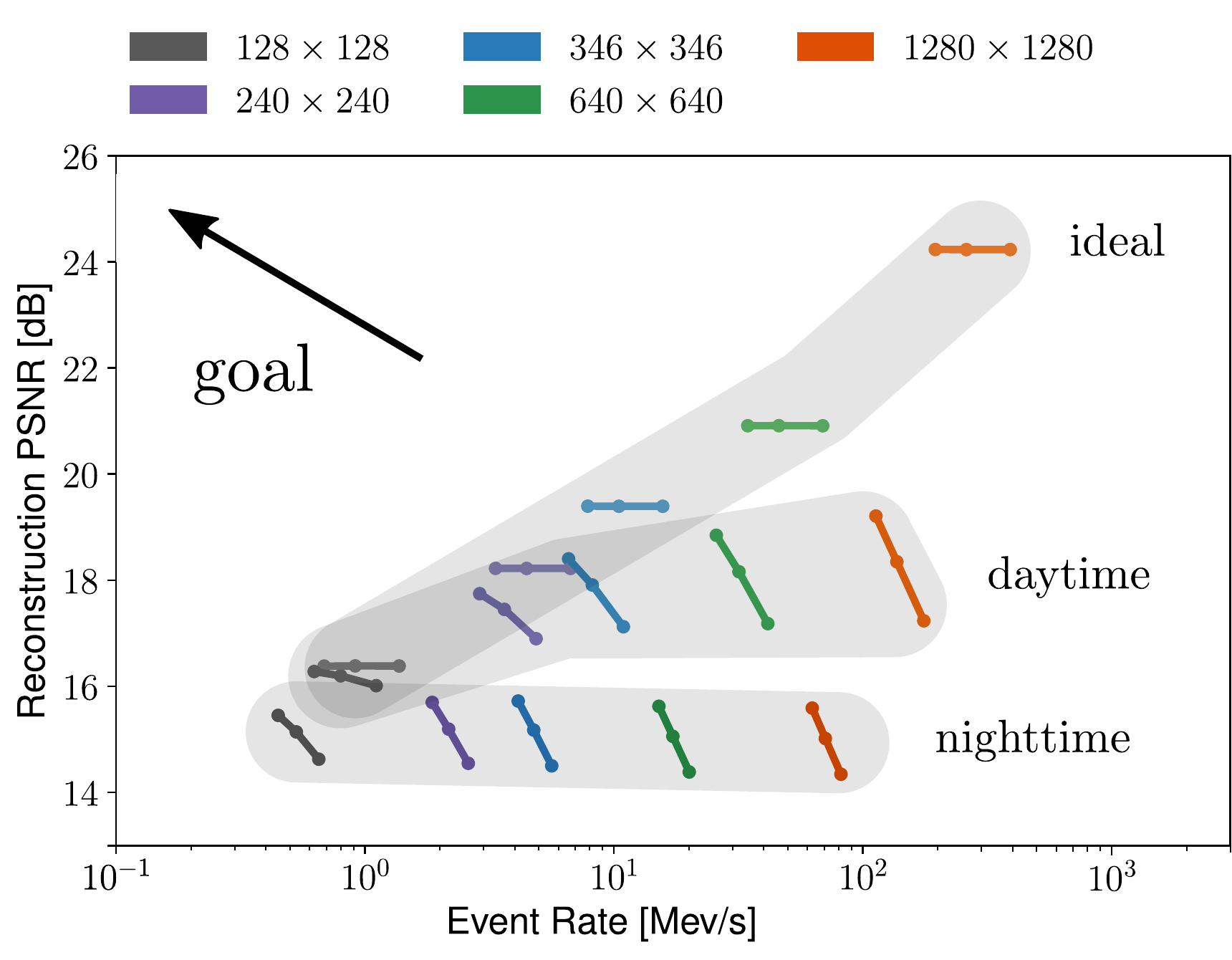}&
        \includegraphics[height=3cm]{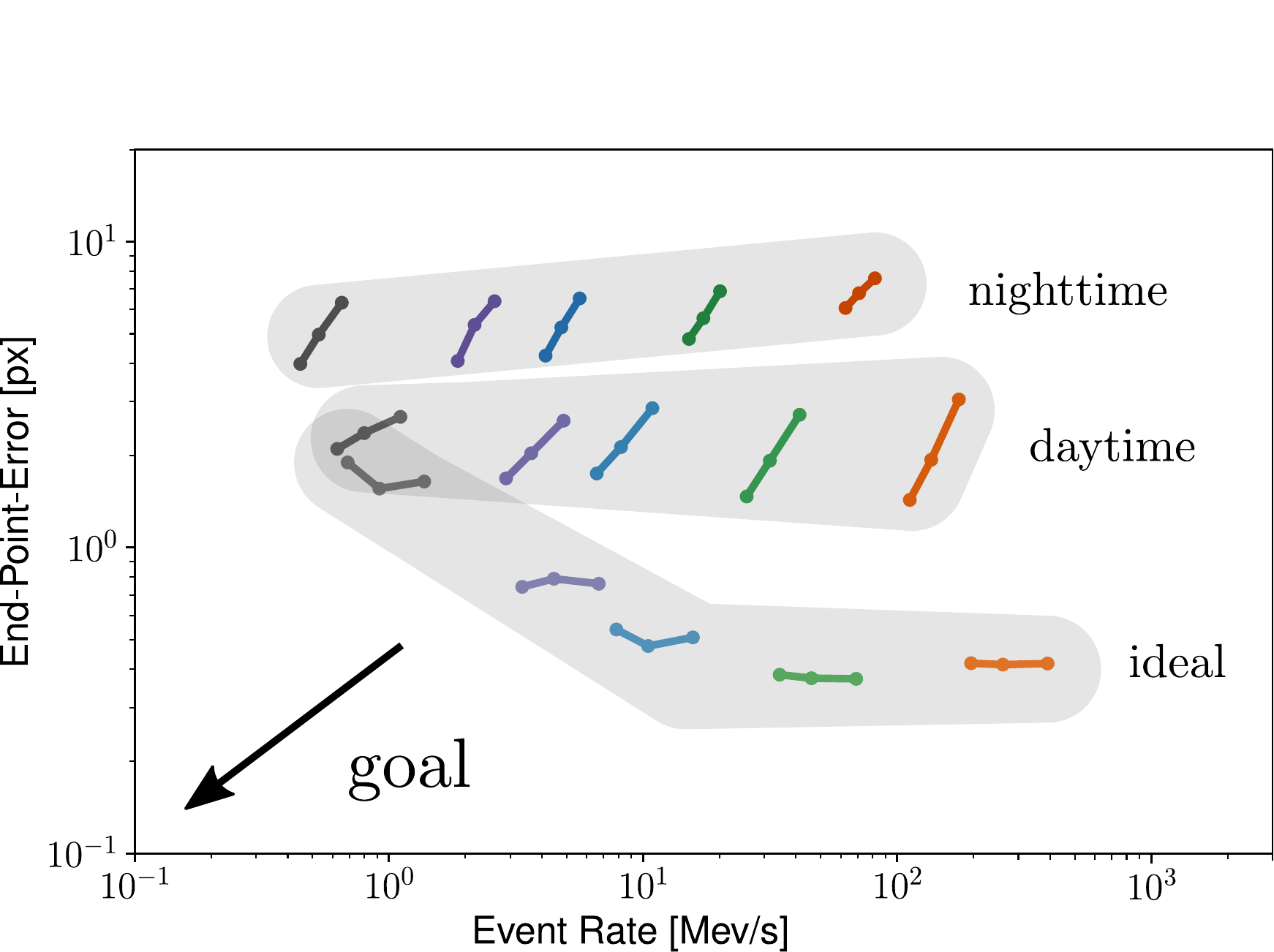}&
        \includegraphics[height=3cm]{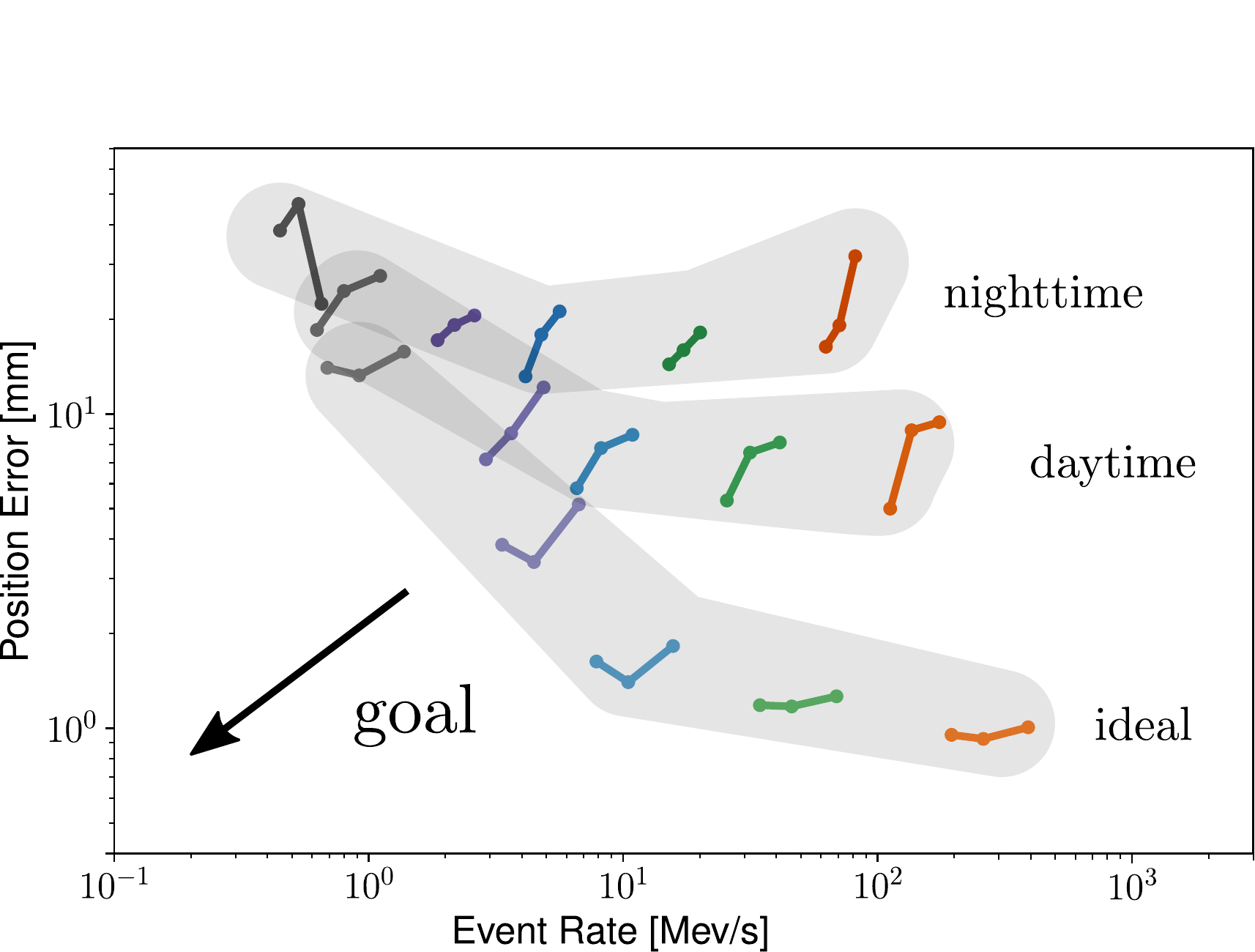}\\
        (a) Image Reconstruction & (b) Optical Flow Estimation &(c) Camera Pose Tracking
    \end{tabular}
    \caption{Each plot compares the event rate and task-performance on the \emph{rocks} sequence in different settings.
Each point corresponds to the median task performance and event rate of a specific setting. Equally colored points correspond  sequences recorded at the same resolution, while connected dots correspond to the same camera trajectory, 
traversed at different speeds (the more right, the faster the sequence, resulting in a higher event rate). 
Finally, each plot is separated into three gray bands which correspond to different cutoff frequency settings. 
These settings correspond with ideal ($f_\text{cutoff}=\infty$), daytime ($f_\text{cutoff}=200$Hz) and nighttime ($f_\text{cutoff}=50$Hz) conditions.
As $f_\text{cutoff}$ decreases, so does task performance. For each task, condition (ideal/daytime/nighttime), and camera velocity, the ideal camera is the one that is closest to the ``goal", maximizing task performance, while minimizing event rate.}\label{fig:tasks_plots}\vspace{-2ex}
\end{figure*}

\section{Experiments}
First, we evaluate the introduced baselines and state-of-the-art methods on a synthetic multi-scale dataset, where we show a tradeoff between resolution, task performance, and event rate. 
Finally, we show similar conclusions for real data.

\subsection{Synthetic Data:}
\label{sec:syn_data}
We use the event camera simulator in \cite{Hu21cvprw} to generate a synthetic multi-scale event camera dataset. 
Each sequence features a random camera trajectory in a planar scene with varying textures. We select resolutions  $r_i=(w,w)$ with $w~\in~\{128, 240, 346, 640, 1280\}$, 
to ensure the same field of view across resolutions. 
The selected widths were selected following commonly used resolutions in Tab. \ref{tab:event_cameras}. 
We use the event camera simulator in \cite{Hu21cvprw} to generate events for each resolution.
In addition to varying the resolution, we also select cutoff frequency $f_{\text{cutoff}}\in \{\infty, 200, 50\}$ Hz to simulate ideal, daytime, and nighttime settings.
These values represent realistic settings and were calibrated in \cite{Hu21cvprw}.
We also vary the camera speed along a given trajectory by scaling it by a factor $s\in \{1, 1.3, 2\}$.
We choose a realistic contrast threshold of $0.2$ for all experiments.

\subsubsection{Image Reconstruction}
\label{sec:image_reconstruction}
For the image reconstruction task, we use the method from Sec. \ref{sec:tasks}.
We use a 50 ms sliding window of events at resolution $r\times r$ with $r \leq 1280$ together with the left keyframe at resolution $1280\times 1280$ to generate predictions of the right keyframe according to Eq.\eqref{eq:event_quality}.
The window is reduced appropriately for higher camera speeds.
We compute the peak signal-to-noise ratio (PSNR) between the predicted and ground truth right keyframe and the event rate, reporting their median across all windows in Fig. \ref{fig:tasks_plots} (a).
We summarized these results in Tab. \ref{tab:all_tasks_syn} (left), and report the full table in the appendix.\\

\begin{table*}[]
    \begin{center}
    \resizebox{0.9\linewidth}{!}{%
    \begin{tabular}{lc|ccc|ccc|ccc}
    \hline
                                     & \textbf{Task}          & \multicolumn{3}{c|}{\textbf{Image Reconstruction}}  & \multicolumn{3}{c|}{\textbf{Optical Flow Estimation}} & \multicolumn{3}{c}{\textbf{Camera Pose Tracking}} \\
                                     & \textbf{Metric}        & \multicolumn{3}{c|}{\textbf{PSNR [dB] $\uparrow$}}             & \multicolumn{3}{c|}{\textbf{RNEPE [px] $\downarrow$}}                & \multicolumn{3}{c}{\textbf{Position Error [mm] $\downarrow$}}  \\ \hline
    \textbf{Dataset}                 & \textbf{Setting}       & \textbf{ideal}  & \textbf{daytime}  & \textbf{nighttime}  & \textbf{ideal}   & \textbf{daytime}   & \textbf{nighttime}  & \textbf{ideal}  & \textbf{daytime} & \textbf{nighttime} \\
                                     & $f_\text{cutoff}$ [Hz] & $\infty$        & 200             & 50              & $\infty$         & 200              & 50              & $\infty$        & 200            & 50             \\ \hline
    \multirow{5}{*}{\textit{carpet}} & $128\times 128$        & 15.389          & 15.343          & 14.892          & 2.09             & 2.11             & \textbf{3.24}      & 37.5            & 37.3           & 29.5           \\
                                     & $240\times 240$        & 17.122          & 16.954          & 15.474          & 0.91             & 0.75             & 3.69      & 17.1            & 21.6           & 33.0           \\
                                     & $346\times 346$        & 18.373          & 17.918          & 15.688          & 0.58             & \textbf{0.65}             & 3.60      & 2.34            & 5.82           & 30.9           \\
                                     & $640\times 640$        & 20.375          & 19.11           & \textbf{15.719} & \textbf{0.36}             & 0.88             & 3.51      & 0.888           & 2.53           & 11.0           \\
                                     & $1280\times 1280$      & \textbf{24.093} & \textbf{20.362} & 15.699          & 0.48             & 0.94             & 3.48      & \textbf{0.617}  & \textbf{2.21}  & \textbf{6.39}  \\ \hline
    \multirow{5}{*}{\textit{rocks}}  & $128\times 128$        & 16.381          & 16.279          & 15.453          & 1.90             & 2.10             & \textbf{3.98}  & 14.0            & 18.5           & 38.4           \\
                                     & $240\times 240$        & 18.224          & 17.747          & 15.697          & 0.74             & 1.68             & 4.07  & 3.84            & 7.17           & 17.2           \\
                                     & $346\times 346$        & 19.397          & 18.403          & \textbf{15.722} & 0.53             & 1.74             & 4.24  & 1.63            & 5.81           & \textbf{13.2}  \\
                                     & $640\times 640$        & 20.913          & 18.848          & 15.624          & \textbf{0.38}             & 1.47             & 4.81  & 1.18            & 5.3            & 14.4           \\
                                     & $1280\times 1280$      & \textbf{24.236} & \textbf{19.211} & 15.590          & 0.42             & \textbf{1.43}             & 6.07  & \textbf{0.953}  & \textbf{5.0}   & 16.4           \\ \hline
    \multirow{5}{*}{\textit{room}}   & $128\times 128$        & 24.786          & 24.44           & 22.282          & 2.36             & 2.21             & 8.54  & -               & -              & -              \\
                                     & $240\times 240$        & 27.764          & 26.687          & \textbf{22.379} & 1.26             & 1.76             & 6.96  & -               & -              & -              \\
                                     & $346\times 346$        & 29.853          & 28.015          & 22.372          & 1.03             & 1.84             & 6.68  & -               & -              & -              \\
                                     & $640\times 640$        & 31.695          & 28.885          & 22.229          & \textbf{0.96}    & \textbf{1.70}    & 6.32  & -               & -              & -              \\
                                     & $1280\times 1280$      & \textbf{34.184} & \textbf{29.844} & 22.167          & 1.33             & 1.95             & \textbf{6.23}    & -               & -              & -              \\ \hline
    \end{tabular}}    
    \end{center}\vspace{-2ex}
    \caption{Task performance on the \emph{carpet}, \emph{rocks} and \emph{room} sequences. 
    We compare differing spatial resolutions on the tasks of image reconstruction, optical flow estimation, and camera pose tracking. We evaluate each setting in ideal,
    daytime, and nighttime conditions. In each column, we highlight the resolution with the best task performance for that setting. 
    In ideal settings performance is most often maximized at the highest resolutions. In daytime settings, the highest resolutions are beneficial in most cases but are outperformed by lower resolution cameras, especially for the task of optical flow estimation. 
    Finally, in nighttime settings, lower resolution cameras almost always perform best. We omit \emph{room} for pose tracking, due to its non-planarity.}\label{tab:all_tasks_syn}  
\end{table*}

\textbf{Discussion}
In ideal settings ($f=\infty$), the highest resolution event cameras achieve the highest performance, 3.18 dB on average higher than the next highest resolutions, but require a roughly 8 times higher event rate.
The same holds for daytime settings ($f=200$ Hz), but this time with a significantly lower margin of 0.86 dB.%
Interestingly, in this setting, we observe that task performance degrades more at higher event camera resolution than at lower resolution.
The event rate is also reduced, but there remains an approximate factor of 6 between resolutions $1280\times 1280$ and $640\times 640$.
Finally, in nighttime settings, ($f=50$ Hz) we observe a peak at resolutions $240\times240$, $346\times346$ and $640\times640$.%
~This shows that for image reconstruction, high-resolution cameras are outperformed by lower resolution cameras, especially in nighttime conditions.\\

\textbf{Comparison with other Methods:}
We validate these conclusions, by evaluating the image reconstruction performance of the learning-based method E2VID\cite{Rebecq19pami} and model-based method HF\cite{Scheerlinck18accv}, following the protocol described in \cite{Rebecq19pami}. 
We report the image reconstruction performance of the two methods in Tab.~\ref{tab:image_recons_sota} (left).
Similar to Tab.~\ref{tab:all_tasks_syn}, for both methods lower resolution cameras become more attractive in nighttime settings.\\ 

\subsubsection{Optical Flow}
\label{sec:optical_flow}
Here we use the method introduced in Sec. \ref{sec:tasks}. 
We extract 20 features from a target image and reproject them into frames spaced at 0.5 seconds in time. 
For each frame, we extract a spatio-temporal volume of events around each feature of size $61\times 61$ and time window $\Delta T=50$ms. 
We adapt the size and length of this window based on the resolution and speed of the camera, using the stated values for the highest resolution and slowest camera speed.
We select $L_l$ at the same resolution as the events and located at the beginning of the spatio-temporal window, and estimate optical flow by minimizing the objective in Eq. \eqref{eq:photometric_constraint_single_event}, using a first-order method. 
We initialize the flow to the ground truth flow and perturb it with Gaussian noise with $\sigma$=3. At lower resolutions we reduce $\sigma$ to create comparable conditions. 
We compare estimated and ground truth flow using a resolution independent normalized end-point-error (RNEPE)
\begin{equation}
\text{RNEPE}=\frac{H_{\text{max}}}{H}\left\Vert\hat{\textbf{v}}-\textbf{v}\right\Vert_2. 
\end{equation}
Where $\hat{\textbf{v}}$ and $\textbf{v}$ are estimated and ground truth optical flow respectively, $H$ is the sensor height and $H_\text{max}=1280$.
We report the RNEPE in Fig.~\ref{fig:scaling_law} (b), summarize results in Tab.~\ref{tab:all_tasks_syn} (middle), and include the full table in the appendix.\\
\begin{table}[t!]
    \centering
    \resizebox{1\linewidth}{!}{%
    \begin{tabular}{c|cccccc|cccccc}
        \hline
        \multicolumn{1}{l|}{\textbf{Task}} &
          \multicolumn{6}{c|}{\textbf{Image Reconstruction}} &
          \multicolumn{6}{c}{\textbf{Optical Flow}} \\ 
          \multicolumn{1}{l|}{\textbf{Metric}}&\multicolumn{6}{c|}{\textbf{PSNR [dB]}$\uparrow$} &
          \multicolumn{6}{c}{\textbf{RNEPE [px]}$\downarrow$} \\\hline
        \multicolumn{1}{l|}{\textbf{Method}} &
          \multicolumn{3}{c|}{E2VID\cite{Rebecq19pami}} &
          \multicolumn{3}{c|}{HF\cite{Scheerlinck18accv}} &
          \multicolumn{3}{c|}{CM\cite{Gallego18cvpr}} &
          \multicolumn{3}{c}{E-RAFT\cite{Gehrig21threedv}} \\ \hline
        \multicolumn{1}{l|}{$f_{\text{cutoff}}$ [Hz]} &
          $\infty$ &
          200 &
          \multicolumn{1}{c|}{50} &
          $\infty$ &
          200 &
          50 &
          $\infty$ &
          200 &
          \multicolumn{1}{c|}{50} &
          $\infty$ &
          200 &
          50 \\ \hline
        $128\times 128$ &
          11.53 &
          11.54 &
          \multicolumn{1}{c|}{11.39} &
          11.14 &
          11.15 &
          10.97 &
          5.06 &
          4.89 &
          \multicolumn{1}{c|}{4.76} &
          2.44 &
          2.32 &
          2.56 \\
        $240\times 240$ &
          12.19 &
          12.17 &
          \multicolumn{1}{c|}{\textbf{11.63}} &
          11.57 &
          11.58 &
          \textbf{11.14} &
          2.51 &
          2.59 &
          \multicolumn{1}{c|}{2.57} &
          1.01 &
          1.10 &
          1.59 \\
        $346\times 346$ &
          12.25 &
          12.21 &
          \multicolumn{1}{c|}{11.36} &
          11.61 &
          11.60 &
          10.98 &
          1.74 &
          1.69 &
          \multicolumn{1}{c|}{1.72} &
          0.94 &
          0.93 &
          1.24 \\
        $640\times 640$ &
          12.317 &
          12.03 &
          \multicolumn{1}{c|}{10.61} &
          11.65 &
          11.55 &
          10.43 &
          0.71 &
          0.67 &
          \multicolumn{1}{c|}{0.69} &
          0.79 &
          0.84 &
          0.86 \\
        $1280\times 1280$ &
          \textbf{13.38} &
          \textbf{12.33} &
          \multicolumn{1}{c|}{10.05} &
          \textbf{12.63} &
          \textbf{12.04} &
          9.93 &
          \textbf{0.32} &
          \textbf{0.25} &
          \multicolumn{1}{c|}{\textbf{0.37}} &
          \textbf{0.61} &
          \textbf{0.63} &
          \textbf{0.68} \\ \hline
        \end{tabular}}\caption{Comparison on the \emph{carpet} sequence of state-of-the-art methods\cite{Rebecq19pami} and \cite{Scheerlinck18accv} on image reconstruction and 
        \cite{Gallego18cvpr} and \cite{Gehrig21threedv} on optical flow estimation.}\label{tab:image_recons_sota}
\end{table}

\begin{table}[t!]
    \centering
    \resizebox{0.8\linewidth}{!}{%
    \begin{tabular}{l|cccc|ccc}
        \hline
    \textbf{Task} & \multicolumn{4}{c|}{\textbf{Optical Flow Estimation}} & \multicolumn{3}{c}{\textbf{Camera Pose Tracking}} \\ \hline
    \textbf{Method} & \multicolumn{4}{c|}{CM\cite{Gallego18cvpr}} & \multicolumn{3}{c}{EPPT (ours)} \\ \hline
    Velocity [m/s] & 0.9 & 1.2 & 1.8 & 9 & 0.9 & 1.2 & 1.8 \\ \hline
    $128\times 128$ & 4.76 & 4.78 & 5.17 & 5.30 & 29.5 & \textbf{2.44} & - \\
    $240\times 240$ & 2.57 & 2.52 & 2.85 & 3.92 & 33.0 & 59.6 & \textbf{4.27} \\
    $346\times 346$ & 1.72 & 1.70 & 1.77 & 3.18 & 30.9 & 41.7 & 11.6 \\
    $640\times 640$ & 0.69 & 0.84 & 1.04 & \textbf{3.09} & 11.0 & 15.4 & 102.0 \\
    $1280\times 1280$ & \textbf{0.37} & \textbf{0.45} & \textbf{0.63} & 3.21 & \textbf{6.39} & 11.5 & 24.9 \\ \hline
    \end{tabular}}
    \caption{Task performance of optical flow method CM \cite{Gallego18cvpr} and our Event-based Photometric Pose Tracker (EPPT), in nighttime conditions ($f_\text{cutoff}=50$Hz) depending on the camera velocity. While PPT with high resolution cameras degrades already at 1.8 m/s, CM degrades at much higher speeds of 9m/s.}\label{tab:perf_vs_speed}
\end{table}
\textbf{Discussion}
As before, Fig.~\ref{fig:tasks_plots} (b) suggests, a trade-off between accuracy and event camera resolution, especially in nighttime and daytime conditions, and at high speeds.
Here we observe that already in clean conditions and at high speeds, lower resolution event cameras perform best, and with a significantly lower event rate. 
This conclusion is supported by Tab. \ref{tab:all_tasks_syn} (middle), where lower resolution camera outperform higher resolution ones by between 0.24-2.09 px. 
On \emph{room}, the highest resolution camera outperforms others by a small margin of 0.09, but it features a roughly three times higher event rate (Fig. \ref{fig:tasks_plots} (b)).\\

\textbf{Comparison with other Methods}
Here we compare task performance for model-based method Contrast Maximization \cite{Gallego18cvpr} and learning-based method E-RAFT\cite{Gehrig21threedv}.
Surprisingly, we observe that CM is more robust than EPF at high camera speeds and in nighttime conditions (Tab~\ref{tab:image_recons_sota}) (right). 
However, at higher speeds of 9 m/s, the $640\times 640$ camera outperforms $1280\times1280$ by 0.2 px (Tab. \ref{tab:perf_vs_speed}).
\rev{Interestingly, this suggests that estimators which optimize contrast instead of phtometric consistency are more robust, 
especially at high speeds and in low light.}
Finally, E-RAFT always benefits from high resolution and does not degrade harshly in noisier conditions. 
Since E-RAFT was trained on real data both in daytime and nighttime conditions, it most likely generalizes to these conditions.
\rev{This indicates another important lesson that can be drawn from these experiments: We may overcome the limitations of noise at high resolutions,
by adopting a learning-base approach, and training on noisy night-time data, robustifying the network against these error sources.
Note, that noisy training-data is the key here, since learning-based method E2VID exhibits the opposite trend, and in fact E2VID is trained with 
synthetic noise-less event data.}

\subsubsection{Camera Pose Tracking}
\label{sec:pose_tracking}
Finally, we evaluate our pose tracker EPPT (Sec.~\ref{sec:tasks}).
In steps of 0.5 seconds, we initialize a spline with 10 control poses spaced at 2 ms with ground truth and then perturb the translation component with Gaussian noise with $\sigma=20$cm.
We minimize Eq. \eqref{eq:pose_tracking} to recover the original poses, repeating this process 20 times. At each time we compute the mean translation error between ground truth and estimated spline.
We compute the median error over samples which we report in Fig. \ref{fig:tasks_plots} (c), summarize in 
Tab. \ref{tab:all_tasks_syn} (right). The full table is in the appendix. We omit \textit{room}, as it is not a planar scene.\\

\textbf{Discussion}
Similar to previous tasks, pose tracking degrades in nighttime conditions and at high speeds (Fig.\ref{fig:tasks_plots}). Performance degrades strongly at high resolution, especially in nighttime conditions, where the error increases from 0.953 to 16.4 mm for resolutions $1280\times 1280$ on the \emph{rocks} dataset. High resolutions perform well on the \emph{carpet} sequence, even in nighttime conditions, but break down as soon as high speeds are introduced (Tab. \ref{tab:perf_vs_speed}).

\subsection{Real Data}
\label{sec:real_data}
\begin{figure*}[t!]
    \centering
    \begin{tabular}{cccc}
        \includegraphics[height=0.15\linewidth]{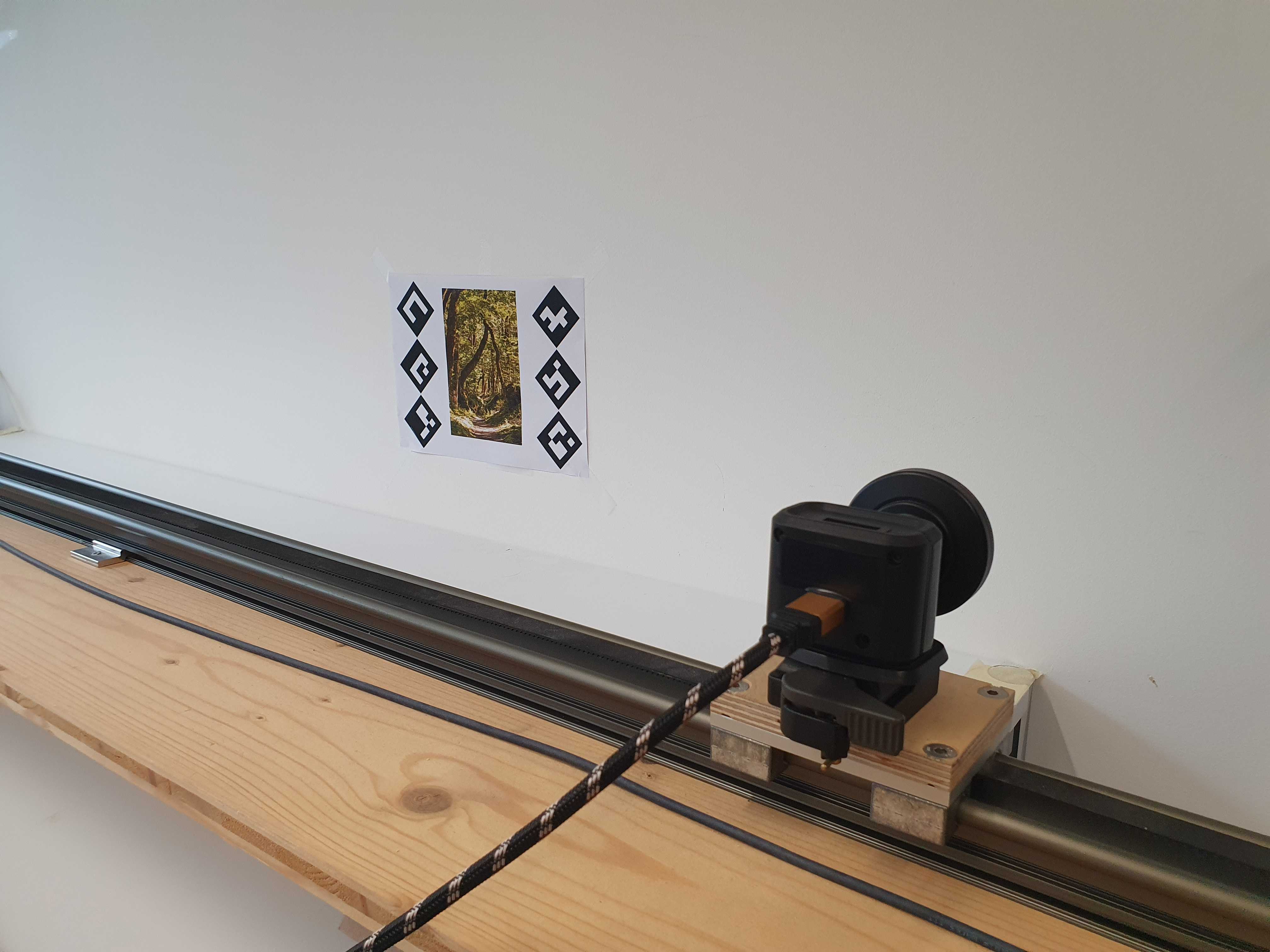}&
        \includegraphics[height=0.15\textwidth]{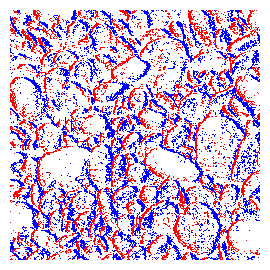}&
        \includegraphics[height=0.15\textwidth]{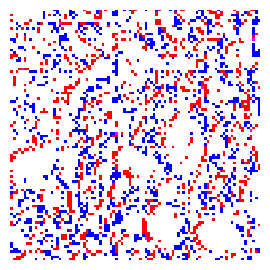}&
        \includegraphics[height=0.15\textwidth]{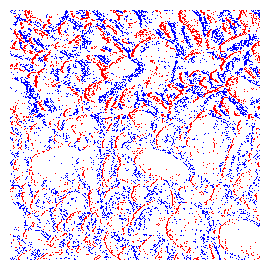}\\
        (a) linear slider& (b) HR, low speed& (c) LR, low speed& (d) HR, high speed
    \end{tabular}
    \caption{We collect a multi-scale event-camera dataset using a linear slider (a), placed front-parallel at a depth of 20 cm, 40 cm, and 60 cm from a wall,
    with a template image with known dimensions. We record events at speeds of 0.2 m/s, 0.4 m/s and 0.6 m/s, resulting 
    in events at different resolutions (a,c) and camera speeds (b,d). These include high-resolution (HR) events and low-resolution (LR) events.
    }\label{fig:linear_slider_setup}
\end{figure*}

Here we show how the conclusions drawn from the experiments on synthetic data transfer to the real world. 
To this end, we propose an experimental setup to record event camera data at different resolutions, observing the same scene. 
We evaluate image reconstruction, optical flow estimation, and pose tracking on the recorded dataset.

We use a Prophesee Gen4 event camera\cite{Finateu20isscc} with a resolution of $1280\times 720$ installed on a motorized linear slider (Fig. \ref{fig:linear_slider_setup} a) in front of a planar wall, featuring a picture of a template image with ArUco markers\cite{Garrido14PR}.
With this setup, we record events at three depths $d\in \{20, 40, 60\}$~cm from the wall, and at three different camera speeds $V\in \{0.2, 0.4, 0.6\}$~m/s. We leverage the slider and markers to recover the camera pose with respect to the printed image. We use this pose to crop events to a common field of view, resulting in events at different resolutions (Fig.\ref{fig:linear_slider_setup} (b), (c)) and speeds (Fig.\ref{fig:linear_slider_setup} (d)). 
More details on how the camera pose was recovered are provided in the appendix.

\subsubsection{Image Reconstruction}
\begin{figure}[t!]
    \centering
    \begin{tabular}{cccc}
        \includegraphics[height=1.7cm]{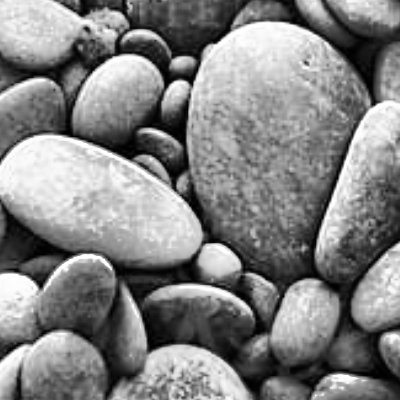}&
        \includegraphics[height=1.7cm]{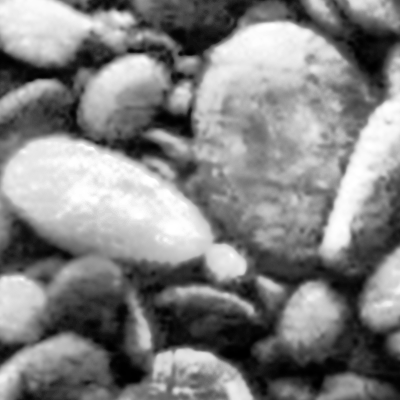}&
        \includegraphics[height=1.7cm]{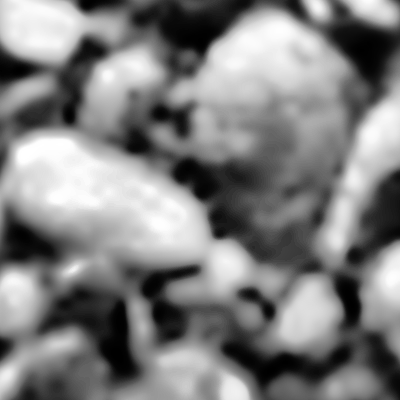}&
        \includegraphics[height=1.7cm]{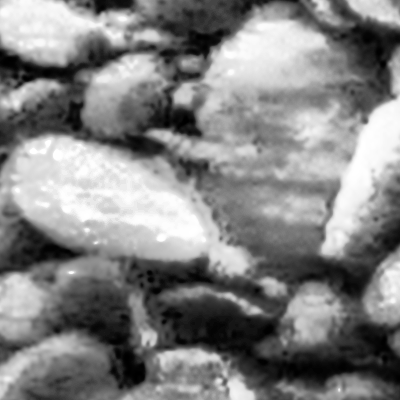}\\
        (a) GT & (b) high res. & (c) low res. & (d) high speed
    \end{tabular}
    \caption{Comparison of image reconstructions by E2VID\cite{Rebecq19pami} in different conditions. 
    Image reconstructions at low-resolution show blurring artifacts (c). Reconstructions at high resolution and high speed
    show significant ghosting (d).}\label{fig:image_recons_qualitative}
\end{figure}

\begin{table*}[]
    \centering
    \resizebox{\linewidth}{!}{%
\begin{tabular}{lc|cccccc|ccccccccc|ccc}
\hline
                                        & \textbf{Task}           & \multicolumn{6}{c|}{\textbf{Image Reconstruction}}                                                                             & \multicolumn{9}{c|}{\textbf{Optical Flow Estimation}}                                                                                                                                                                                & \multicolumn{3}{c}{\textbf{Pose Tracking}}             \\
                                        & \textbf{Metric}         & \multicolumn{6}{c|}{\textbf{PSNR [dB] $\uparrow$}}                                                                             & \multicolumn{9}{c|}{\textbf{RNEPE [px]} $\downarrow$}                                                                                                                                                               & \multicolumn{3}{c}{\textbf{Pos. Error [mm] $\downarrow$}} \\ \hline
                                        & \textbf{Method}         & \multicolumn{3}{c|}{E2VID\cite{Rebecq19pami}}                                      & \multicolumn{3}{c|}{HF\cite{Scheerlinck18accv}}                    & \multicolumn{3}{c|}{EPF (ours)}                                                  & \multicolumn{3}{c|}{CM\cite{Gallego18cvpr}}                                                  & \multicolumn{3}{c|}{E-RAFT\cite{Gehrig21threedv}}                         & \multicolumn{3}{c}{EPPT}                               \\
\textbf{Dataset}                      & \textbf{Velocity [m/s]} & 0.2             & 0.4             & \multicolumn{1}{c|}{0.6}             & 0.2             & 0.4             & 0.6             & 0.2                & 0.4                & \multicolumn{1}{c|}{0.6}                & 0.2                & 0.4                & \multicolumn{1}{c|}{0.6}                & 0.2                & 0.4                & 0.6                & 0.2                 & 0.4                & 0.6                \\ \hline
\multirow{3}{*}{\textit{forest}}      & $442\times 249$         & 12.79          & 12.72          & \multicolumn{1}{c|}{12.54}          & 12.24          & 12.21          & 12.03          & 2.44 & 2.84 & \multicolumn{1}{c|}{3.10} & 1.96 & 2.24 & \multicolumn{1}{c|}{2.42} & 0.64          & 0.64          & 0.75          & 17.47               & 24.44              & 64.97              \\
                                        & $649\times 365$       & 13.56          & \textbf{13.50} & \multicolumn{1}{c|}{\textbf{13.25}} & 12.91          & \textbf{12.85} & \textbf{12.55} & 1.36 & 1.81 & \multicolumn{1}{c|}{\textbf{2.27}} & 1.05 & 1.41 & \multicolumn{1}{c|}{1.60} & 0.33          & 0.36          & 0.45          & 4.08                & 19.96              & \textbf{61.53}     \\
                                        & $1190\times 669$      & \textbf{14.09} & 13.33          & \multicolumn{1}{c|}{12.67}          & \textbf{13.37} & 12.64          & 11.89          & \textbf{1.17} & \textbf{1.77} & \multicolumn{1}{c|}{2.32} & \textbf{0.39} & \textbf{0.53} & \multicolumn{1}{c|}{\textbf{0.79}} & \textbf{0.31} & \textbf{0.34} & \textbf{0.37} & \textbf{2.72}       & \textbf{14.58}     & 75.55              \\ \hline
\multirow{3}{*}{\textit{pebbles}}     & $451\times 254$         & 14.37          & 13.79          & \multicolumn{1}{c|}{13.15}          & 13.62          & 13.06          & 12.42          & 1.23 & 1.67 & \multicolumn{1}{c|}{2.20} & 1.36 & 1.61 & \multicolumn{1}{c|}{2.02} & 0.52          & 0.59          & 0.70          & 5.80                & 27.13              & 79.42              \\
                                      & $652\times 367$         & \textbf{14.82} & \textbf{14.24} & \multicolumn{1}{c|}{13.65}          & \textbf{13.82} & \textbf{13.27} & \textbf{12.72} & 0.90 & 1.40 & \multicolumn{1}{c|}{\textbf{1.76}} & 0.73 & 0.947 & \multicolumn{1}{c|}{1.16} & 0.34          & 0.36          & 0.43          & 3.97                & \textbf{11.34}     & \textbf{49.65}     \\
                                      & $1215\times 683$        & 14.41          & 14.16          & \multicolumn{1}{c|}{\textbf{13.67}} & 13.395          & 13.00          & 12.32          & \textbf{0.84} & \textbf{1.29} & \multicolumn{1}{c|}{1.92} & \textbf{0.31} & \textbf{0.39} & \multicolumn{1}{c|}{\textbf{0.54}} & \textbf{0.34} & \textbf{0.35} & \textbf{0.38} & \textbf{3.21}       & 26.98              & 97.08              \\ \hline
\multirow{3}{*}{\textit{split\_rocks}}& $442 \times 249$        & 12.21           & 12.31           & \multicolumn{1}{c|}{\textbf{12.16}}  & 11.56           & 11.69           & \textbf{11.58}  & 2.16 & 2.59 & \multicolumn{1}{c|}{3.01} & 1.52 & 1.80 & \multicolumn{1}{c|}{2.23} & 0.49          & 0.53          & 0.61          & 11.97               & 41.33              & 50.25              \\
                                      & $640\times 360$         & 12.73           & 12.37           & \multicolumn{1}{c|}{12.06}           & 12.02           & 11.81           & 11.50           & 1.88 & 2.36 & \multicolumn{1}{c|}{\textbf{2.64}} & 0.91 & 1.31 & \multicolumn{1}{c|}{1.66} & 0.36          & 0.38          & 0.36          & 3.74                & 12.00              & \textbf{47.11}     \\
                                      & $1193\times 671$        & \textbf{12.76}  & \textbf{12.52}  & \multicolumn{1}{c|}{11.24}           & \textbf{12.09}  & \textbf{11.89}  & 11.56           & \textbf{1.87} & \textbf{2.39} & \multicolumn{1}{c|}{2.64} & \textbf{0.45} & \textbf{0.76} & \multicolumn{1}{c|}{\textbf{0.96}} & \textbf{0.18} & \textbf{0.20} & \textbf{0.23} & \textbf{1.708}      & \textbf{11.11}     & 78.91              \\ \hline
\end{tabular}}
\caption{Comparison of image reconstruction, optical flow estimation and camera pose tracking performance on the \emph{forest}, \emph{pebbles} and \emph{split\_rocks}, sequences. 
While at slow speeds, the highest resolutions are almost always best, at higher speeds lower resolutions outperform the highest. We see that across tasks a resolution of ~$640\times 360$ offers better task performance than the full resolution.}\label{tab:all_tasks_real}
\end{table*}

As before, we generate image reconstructions using E2VID\cite{Rebecq19pami} and HF\cite{Scheerlinck18accv}. 
To ensure equal conditions across resolutions we pass 30 ms events windows to each method, adjusting the window size to the camera speed.
To generate ground truth, we map the template image into the current camera frame, according to the current camera pose. We report the results in Tab. \ref{tab:all_tasks_real} (left). 
As before, at higher speeds higher resolution cameras degrade more than lower resolution cameras. Indeed, 
already at moderate speeds of 0.4 m/s, a resolution of ~$640\times 360$ outperforms higher resolutions by an average of 0.2 dB.

Fig. \ref{fig:image_recons_qualitative} visualizes image reconstruction patches by \cite{Rebecq19pami} in different settings.
At low resolutions, reconstructions show common blurring artefacts (c). 
By contrast at high resolutions and speeds, reconstructions show ghosting (d), which cause significant performance degradations at high speeds. 
This is caused by slow cutoff frequencies in the event pixels, which lead to slower response times.

\subsubsection{Optical Flow and Camera Pose Tracking}
We evaluate the model-based method EPF, described in Sec. \ref{sec:tasks}, CM\cite{Gallego18cvpr} and E-RAFT\cite{Gehrig21threedv} on patch-based optical flow.
For this, we extract corners from the template image and reproject these corners into the current event camera frame. 
By taking differences between projections at consecutive timestamps, we can generate ground truth flow. 
We use this flow to evaluate the methods described. We generate volumes of 30 ms and patch size 61 and report the resolution normalized RNEPE in Tab. \ref{tab:all_tasks_real} (middle).
Similarly, for camera tracking we follow the same procedure as Sec. \ref{sec:pose_tracking}, but this time perturb the splines 
with $\sigma=5$cm. We report the median position error in mm in Tab. \ref{tab:all_tasks_real} (right). For both EPF and EPPT we use image reconstructions from \cite{Rebecq19pami} as reference images.
Similar conclusions hold as for the experiments in simulation. For speeds above 0.4 m/s, pose tracking and optical flow degrade at the highest resolution. 
In both cases, a resolution of ~$640\times 360$ yields better results and lower event rates. While pose tracking performance increases by 31 mm, optical flow performance increases by 0.1 px for EPF. As before, CM and E-RAFT are more robust to high speeds, performing best at the highest resolution. 
\rev{Again this indicates that adopting CM-based objectives can in fact significantly improve the robustness to 
high speed, especially at high resolution. Moreover, adopting a learning-based approach, which is trained on noisy night-time events also helps.}

\section{Conclusion}
\vspace{-1ex}
As event cameras increasingly become the sensor of choice in challenging applications, their development will inexorably trend toward higher resolution sensors. 
However, higher resolution sensors significantly increase the required data bandwidth, while burdening downstream tasks. In this work, we showed, that in addition to these challenges, higher resolution cameras are also more sensitive to temporal effects such as slow pixel response times. We showed across the tasks of image reconstruction, optical flow estimation and camera pose tracking
that high-resolution event cameras do not always give the best task performance. Especially when used in challenging high-speed scenarios and in low light,
lower-resolution sensors often show a better performance while using lower bandwidth. 
It is thus important that additional effort is made to solve these challenges before we can progress to higher resolution sensors. We believe that this work will act as a useful guide for future trends in event camera development.

\section{Acknowledgment}
This work was supported by Huawei, and as a part of NCCR Robotics, a National Centre of Competence in Research, funded by the Swiss National Science Foundation (grant number 51NF40\_185543).

\begin{figure}
\centering
\begin{tabular}{cccc}
     \includegraphics[height=1.8cm]{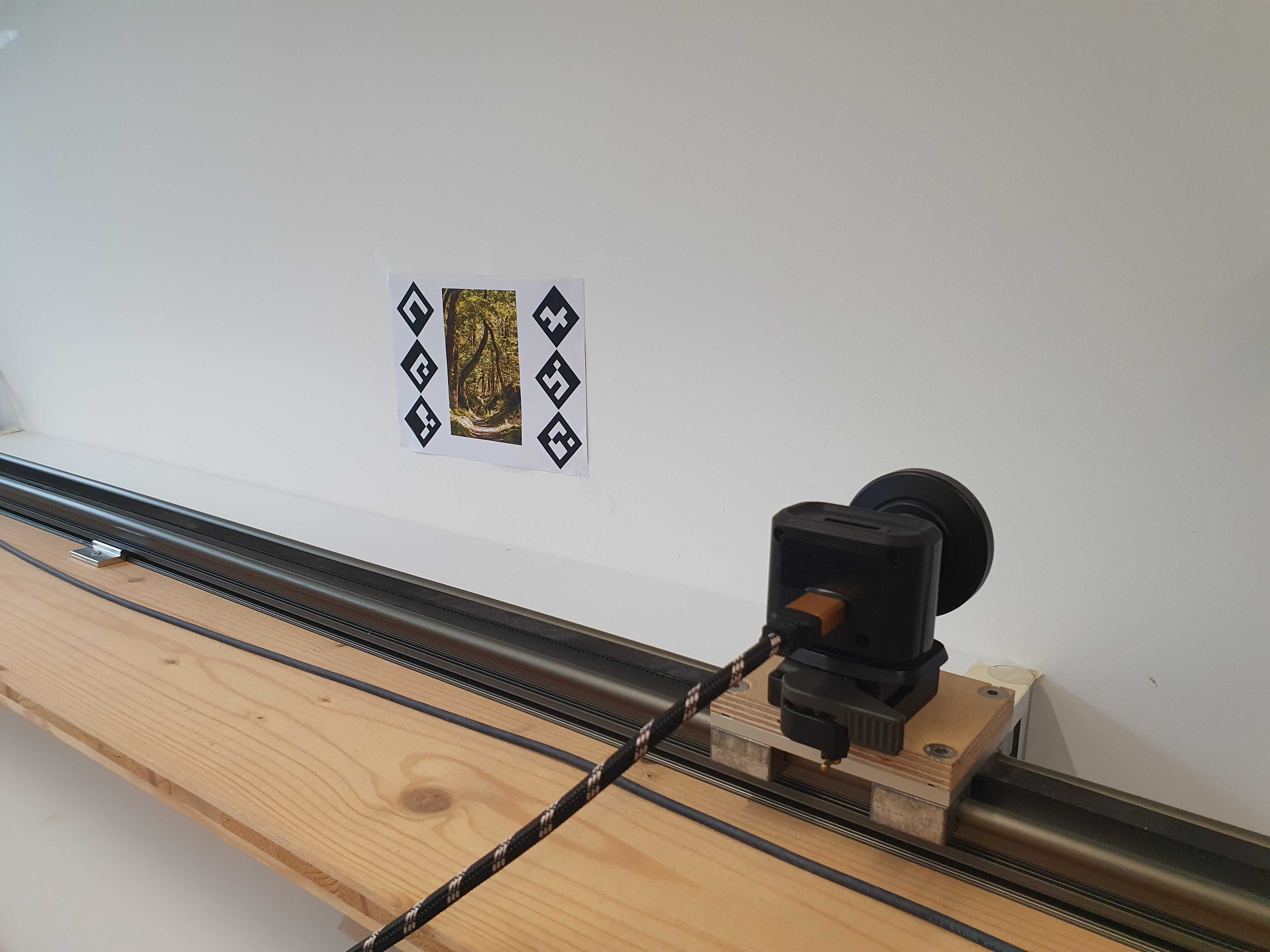}&
     \includegraphics[height=1.8cm]{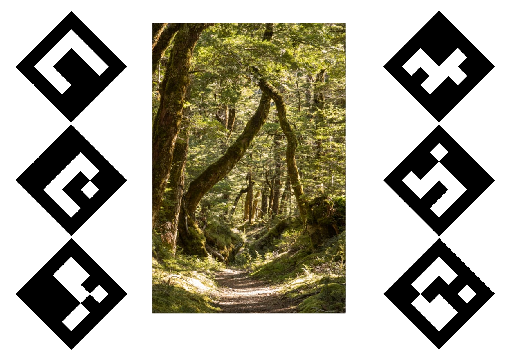}&
     \includegraphics[height=1.8cm,trim=550 50 50 50, clip]{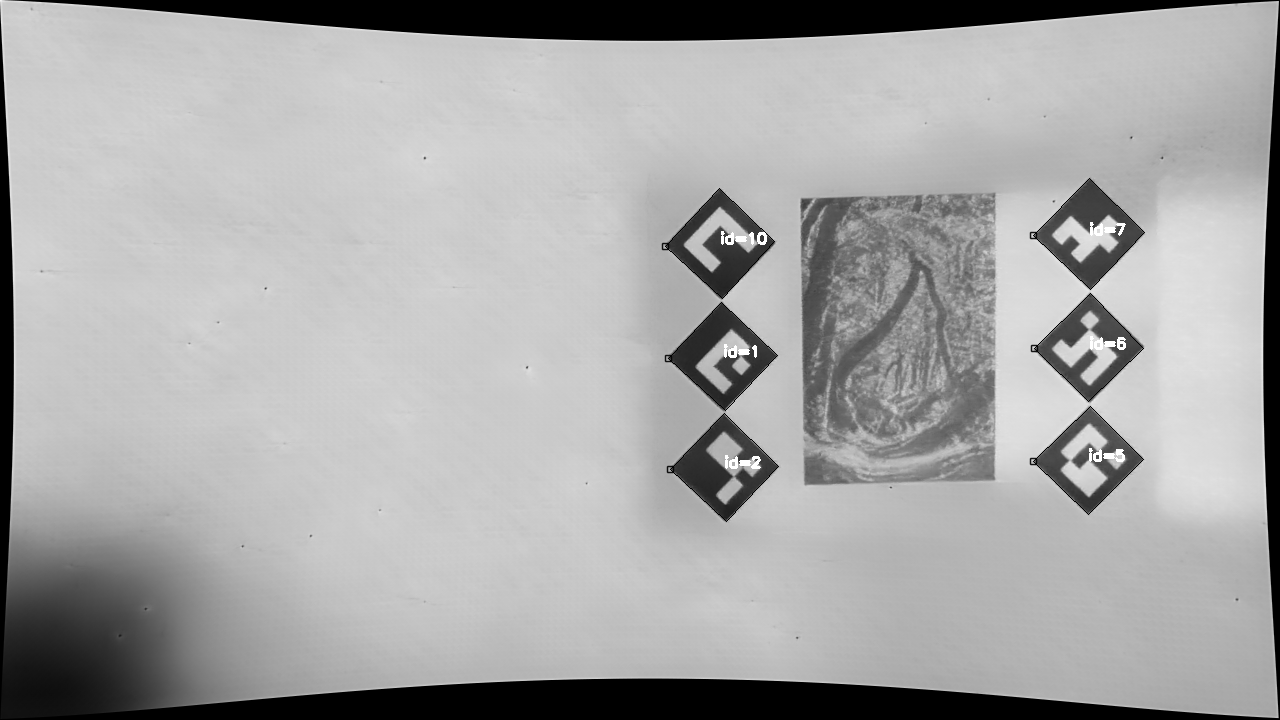}&
     \includegraphics[height=1.8cm]{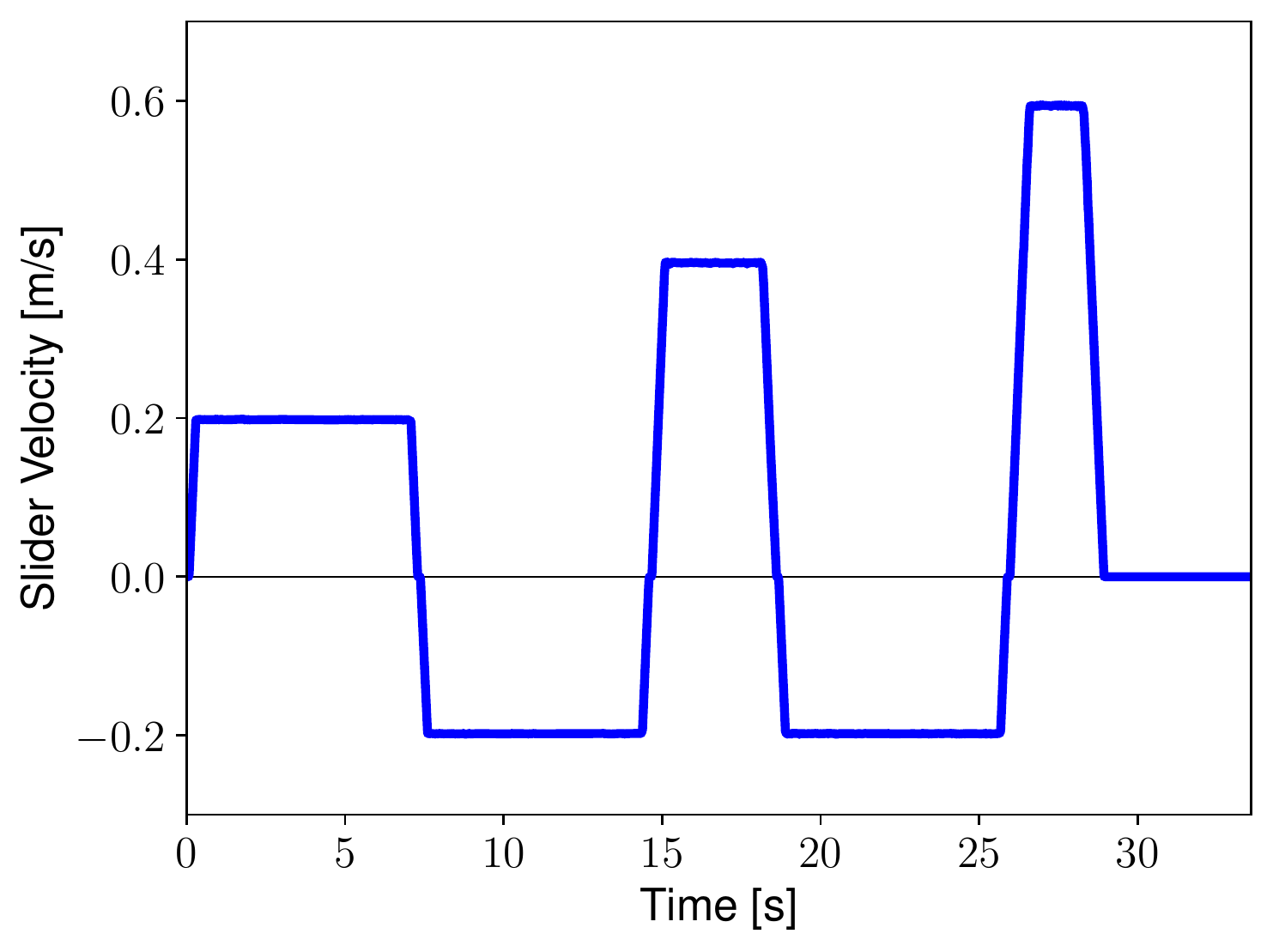}\\
     (a) preview &(b) image template & (c) ArUco detections & (d) speed profile 
\end{tabular}
\caption{Preview of the real world dataset recorded with a linear slider (a). We print an image based on the template in (b) which we stick to the wall. By leveraging the image reconstruction method E2VID\cite{Rebecq19pami}, we reconstruct images from events. We then detect ArUco markers\cite{Garrido14PR}. By completing a velocity profile as in (d) the camera detects sufficient markers to solve a global bundle adjustment problem, that recovers the pose of the camera.}
\label{fig:app:linear_slider}
\end{figure}

\section{Appendix}
In Sec. \ref{sec:app:linear_slider} we add additional details to explain how the real-world multi-scale event dataset was collected. Tables and figures in the main manuscript will have an "M-" prefix. In Sec. \ref{sec:app:all_tasks_all_speeds_all} we report the full table, which was used to generate Fig. M-3, split by tasks, supplementing Tab. M-2 in the main manuscript.  

\subsection{Generation of the Multi-scale Dataset}
\label{sec:app:linear_slider}
We install a Prophesee Gen4 event camera on a motorized linear slider (Fig. M-4 (a)), placed in a front-parallel configuration to a wall, featuring a printed image with ArUco markers\cite{Garrido14PR} surrounding it (Fig.\ref{fig:app:linear_slider} (b)). We leverage the markers and the linear slider to recover the camera pose with respect to the bottom left corners of the printed page, which we designate as the origin. 

We do this in three steps. First we convert events to using E2VID\cite{Rebecq19pami} to generate sharp grayscale images which we use to extract the marker corner positions (Fig.\ref{fig:app:linear_slider} (c)).
The 3-D position of the corners are fully known and fixed, and thus we solve a pose-only global bundle adjustment problem that minimizes the reprojection error between 2-D detections and known landmark positions. The full problem we solve is written as 
\begin{align}
    \{T_i\} = \arg\min_{\{T_i\}} \sum_{i=1}^{N}\sum_{k=1}^{M}v_{ik}\Vert \textbf{x}_{ik}-\pi(\textbf{X}_k,K,T_i)\Vert_2^2,
\end{align}
where $T_i$ denote the camera poses with $i=1,...,N$, $\textbf{X}_k$ denote the fixed landmark positions with $k=1,...,M$ and $K$ captures the camera intrinsics. The parameter $v_{ik}=1$ if the landmark $k$ is in view $i$ and $0$ otherwise. Finally, $\textbf{x}_{ik}$ denote the detection of landmark $k$ in view $i$.
The linear slider additionally provides the $x$-coordinate along the path, which is used to reduce the number of unknowns. 
We thus reduce the number of unknowns to 9 parameters. Six describe the relative orientation and translation between the linear slider base at position $x=0$ and the last three describe the relative rotation of the camera frame with respect to the linear slider frame.
We collect measurements for the bundle adjustment problem by following the camera speed profile in Fig.\ref{fig:app:linear_slider} (d), which drives the slider back and forth a total of three times, at increasing speed. This is done because at high speeds image reconstruction becomes blurry and therefore few detections are made. By inserting a slow-speed segment between each right-moving segment, we ensure that accurate detections can still be made, to constrain bundle adjustment. This allows us to have accurate poses even, where there are no detections. \\

\textbf{Ground Truth Generation:}
Finally, we use the recovered poses for ground truth generation for all three tasks: for image reconstructions, a template image (Fig. \ref{fig:app:linear_slider} (b)) is used, to which all image reconstructions are mapped using homographies. By mapping images at all speeds and resolutions to the same template, we ensure an equal evaluation. For optical flow we use the poses to project points into poses spaced at $\Delta T$, resulting in accurate flow vectors. For camera pose tracking we use the poses themselves.

\subsection{Additional Results in Simulation}
\label{sec:app:all_tasks_all_speeds_all}
Here we summarize additional results in simulation, which for brevity were omitted in Tab. M-2.\\

\textbf{Image Reconstruction} 
In Tab. \ref{tab:app:image_recons_ei} we report the full table, comparing the event integration method, described in Sec. M-3.1 on our synthetic multi-scale dataset. It completes the results in Tab. M-2 (left) and was used to generate Fig. M-3 (a). We report scores for different camera resolutions, cutoff frequencies, and speeds. Moreover, we show in Tab. \ref{tab:app:image_recons_comp} the full table comparing the image reconstruction methods E2VID~\cite{Rebecq19pami} and HF~\cite{Scheerlinck18accv} on the \emph{carpet} sequence. Both tables, show, that the conclusions hold across datasets and methods that higher resolution cameras are outperformed by lower resolution ones, especially in nighttime conditions and at high speeds. \\

\textbf{Optical Flow Estimation}
Similar to image reconstruction, we show the complete table which complements Tab. M-2 (middle) and Fig. M-3 (b). In Tab. \ref{tab:app:optical_flow} we show the optical flow method EPF on all synthetic datasets and for different camera resolutions, speeds and cutoff frequencies. Tab. \ref{tab:app:optical_flow_sota} shows the performance of state-of-the-art flow methods Contrast Maximization (CM)\cite{Gallego18cvpr} and E-RAFT\cite{Gehrig21threedv} on the \emph{carpet} sequence. Tab. \ref{tab:app:optical_flow} indicates the same conclusion of Tab. M-2 (left). The faster the camera and the lower the cutoff frequency (i.e. in nighttime conditions), the worse high-resolution cameras perform. Indeed, already in daytime conditions, sensors below a resolution of $346\times346$ show better performance than those at a resolution of $1280\times 1280$. Tab. \ref{tab:app:optical_flow_sota} reiterates the conclusion made in the main text: For EPF fast camera motion and nighttime conditions lead to significant degradation of the optical flow method at high resolutions. Instead, using lower resolution sensors is better. In nighttime conditions, all resolutions perform poorly, leading to noisy values. CM remains stable at high resolutions, high speeds, and low cutoff frequencies. However, at speeds of 9 m/s in night-time conditions, CM at high resolution is outperformed by a sensor of resolution $640\times 640$ by 0.12 px. E-RAFT at the highest resolution remains stable at all speeds but shows significant degradation when going to high speeds and nighttime conditions. We believe that this is because E-RAFT trains on low-light events and can thus generalize better to this setting. \\

\textbf{Camera Pose Tracking}
Also for camera pose tracking, we show the complete table, complementing Tab. M-2 (right) and Fig. M-3. (c). 
Here we also report the position error for three different camera speeds. We see that especially in nighttime conditions and at high speeds, pose tracking using high-resolution sensors is outperformed by higher resolution sensors.

\begin{table*}
    \centering
    \resizebox{1\linewidth}{!}{%
    \begin{tabular}{lc|ccc|ccc|ccc}
        \hline
        \textbf{Dataset} & \textbf{Setting}  & \multicolumn{3}{c|}{\textbf{ideal} }                      & \multicolumn{3}{c|}{\textbf{daytime}}                         & \multicolumn{3}{c}{\textbf{nighttime}}                           \\ 
        &$f_\text{cutoff}$ [Hz]&\multicolumn{3}{c|}{$\infty$}&\multicolumn{3}{c|}{200}&\multicolumn{3}{c}{50}\\\hline
                     & Velocity [m/s]          & 1.2                     & 1.6                     & 2.4                     & 1.2                     & 1.6                     & 2.4 & 1.2                     & 1.6                     & 2.4 \\ \hline
\multirow{5}{*}{room}                         & $128\times 128$   & \multicolumn{1}{c|}{24.786} & \multicolumn{1}{c|}{24.791} & 24.787 & \multicolumn{1}{c|}{24.44} & \multicolumn{1}{c|}{24.076} & 23.428  & \multicolumn{1}{c|}{22.282} & \multicolumn{1}{c|}{\textbf{21.951}} & \textbf{21.608}  \\
                         & $240\times 240$   & \multicolumn{1}{c|}{27.764} & \multicolumn{1}{c|}{27.768} & 27.769 & \multicolumn{1}{c|}{26.687} & \multicolumn{1}{c|}{25.842} & 24.493  & \multicolumn{1}{c|}{\textbf{22.379}} & \multicolumn{1}{c|}{21.809} & 21.355  \\
                         & $346\times 346$   & \multicolumn{1}{c|}{29.853} & \multicolumn{1}{c|}{29.853} & 29.854 & \multicolumn{1}{c|}{28.015} & \multicolumn{1}{c|}{26.806} & 24.97  & \multicolumn{1}{c|}{22.372} & \multicolumn{1}{c|}{21.751} & 21.278  \\
                         & $640\times 640$   & \multicolumn{1}{c|}{31.695} & \multicolumn{1}{c|}{31.686} & 31.694 & \multicolumn{1}{c|}{28.885} & \multicolumn{1}{c|}{27.218} & 25.007  & \multicolumn{1}{c|}{22.229} & \multicolumn{1}{c|}{21.609} & 21.171  \\
                         & $1280\times 1280$ & \multicolumn{1}{c|}{\textbf{34.184}} & \multicolumn{1}{c|}{\textbf{34.18}} & \textbf{34.179} & \multicolumn{1}{c|}{\textbf{29.844}} & \multicolumn{1}{c|}{\textbf{27.605}} & \textbf{25.081}  & \multicolumn{1}{c|}{22.167} & \multicolumn{1}{c|}{21.548} & 21.123  \\ \hline
                   & Velocity [m/s]          & 0.9                     & 1.2                     & 1.8                      & 0.9                     & 1.2                     & 1.8 & 0.9                     & 1.2                     & 1.8 \\ \hline
\multirow{5}{*}{carpet}                         & $128\times 128$   & \multicolumn{1}{c|}{15.389} & \multicolumn{1}{c|}{15.39} & 15.385 & \multicolumn{1}{c|}{15.343} & \multicolumn{1}{c|}{15.314} & 15.233  & \multicolumn{1}{c|}{14.892} & \multicolumn{1}{c|}{14.65} & 14.283  \\
                         & $240\times 240$   & \multicolumn{1}{c|}{17.122} & \multicolumn{1}{c|}{17.12} & 17.123 & \multicolumn{1}{c|}{16.954} & \multicolumn{1}{c|}{16.785} & 16.414  & \multicolumn{1}{c|}{15.474} & \multicolumn{1}{c|}{14.973} & \textbf{14.326}  \\
                         & $346\times 346$   & \multicolumn{1}{c|}{18.373} & \multicolumn{1}{c|}{18.374} & 18.374 & \multicolumn{1}{c|}{17.918} & \multicolumn{1}{c|}{17.617} & 17.016  & \multicolumn{1}{c|}{15.688} & \multicolumn{1}{c|}{\textbf{15.056}} & 14.313  \\
                         & $640\times 640$   & \multicolumn{1}{c|}{20.375} & \multicolumn{1}{c|}{20.376} & 20.375 & \multicolumn{1}{c|}{19.11} & \multicolumn{1}{c|}{18.47} & 17.464  & \multicolumn{1}{c|}{\textbf{15.719}} & \multicolumn{1}{c|}{14.997} & 14.221  \\
                         & $1280\times 1280$ & \multicolumn{1}{c|}{\textbf{24.093}} & \multicolumn{1}{c|}{\textbf{24.096}} & \textbf{24.101} & \multicolumn{1}{c|}{\textbf{20.362}} & \multicolumn{1}{c|}{\textbf{19.184}} & \textbf{17.731}  & \multicolumn{1}{c|}{15.699} & \multicolumn{1}{c|}{14.949} & 14.171  \\ \hline
                    & Velocity [m/s]         & 0.7                     & 1.0                     & 1.4                      & 0.7                     & 1.0                     & 1.4 & 0.7                     & 1.0                     & 1.4 \\ \hline
        \multirow{5}{*}{rocks}                 & $128\times 128$   & \multicolumn{1}{c|}{16.381} & \multicolumn{1}{c|}{16.385} & 16.384  & \multicolumn{1}{c|}{16.279} & \multicolumn{1}{c|}{16.198} & 16.013 & \multicolumn{1}{c|}{15.453} & \multicolumn{1}{c|}{15.141} & \textbf{14.627}  \\
                         & $240\times 240$   & \multicolumn{1}{c|}{18.224} & \multicolumn{1}{c|}{18.223} & 18.223  & \multicolumn{1}{c|}{17.747} & \multicolumn{1}{c|}{17.45} & 16.899 & \multicolumn{1}{c|}{15.697} & \multicolumn{1}{c|}{\textbf{15.194}} & 14.549  \\
                         & $346\times 346$   & \multicolumn{1}{c|}{19.397} & \multicolumn{1}{c|}{19.395} & 19.395  & \multicolumn{1}{c|}{18.403} & \multicolumn{1}{c|}{17.91} & 17.123 & \multicolumn{1}{c|}{\textbf{15.722}} & \multicolumn{1}{c|}{15.175} & 14.501  \\
                         & $640\times 640$   & \multicolumn{1}{c|}{20.913} & \multicolumn{1}{c|}{20.914} & 20.912  & \multicolumn{1}{c|}{18.848} & \multicolumn{1}{c|}{18.162} & 17.182 & \multicolumn{1}{c|}{15.624} & \multicolumn{1}{c|}{15.057} & 14.385  \\
                         & $1280\times 1280$ & \multicolumn{1}{c|}{\textbf{24.236}} & \multicolumn{1}{c|}{\textbf{24.234}} & \textbf{24.235}  & \multicolumn{1}{c|}{\textbf{19.211}} & \multicolumn{1}{c|}{\textbf{18.349}} & \textbf{17.236} & \multicolumn{1}{c|}{15.59} & \multicolumn{1}{c|}{15.017} & 14.343  \\ \hline
        \end{tabular}}
\caption{Effect of resolution on median image reconstruction accuracy (PSNR [dB]) in ideal ($f=\infty$ Hz), daytime ($f=200$ Hz) and nighttime ($f=50$ Hz) settings, and three camera speeds.
While highest resolution event cameras achieve the best task performance in ideal and clean conditions, they do this with only small margins (~0.1 dB), and at the cost of significant event-rates.
In noisy conditions, lower resolution event cameras such as the $346\time 346$ outperform the highest resolution.}\label{tab:app:image_recons_ei}
\end{table*}

\begin{table*}[]
    \centering
    \resizebox{1\linewidth}{!}{%
    \begin{tabular}{lc|ccc|ccc|ccc}
    \hline
    \textbf{Method} & \textbf{Setting}  & \multicolumn{3}{c|}{\textbf{ideal}}                      & \multicolumn{3}{c|}{\textbf{daytime}}                         & \multicolumn{3}{c}{\textbf{nighttime}}                           \\
    &$f_\text{cutoff}$ [Hz]&\multicolumn{3}{c|}{$\infty$}&\multicolumn{3}{c|}{200}&\multicolumn{3}{c|}{50}\\\hline
               & Velocity [m/s]          & 0.9                     & 1.2                     & 1.8                      & 0.9                     & 1.2                     & 1.8 & 0.9                     & 1.2                     & 1.8 \\ \hline
    \multirow{5}{*}{EI (ours)}                     & $128\times 128$   & \multicolumn{1}{c|}{15.389} & \multicolumn{1}{c|}{15.39} & 15.385 & \multicolumn{1}{c|}{15.343} & \multicolumn{1}{c|}{15.314} & 15.233  & \multicolumn{1}{c|}{14.892} & \multicolumn{1}{c|}{14.65} & 14.283  \\
                         & $240\times 240$   & \multicolumn{1}{c|}{17.122} & \multicolumn{1}{c|}{17.12} & 17.123 & \multicolumn{1}{c|}{16.954} & \multicolumn{1}{c|}{16.785} & 16.414  & \multicolumn{1}{c|}{15.474} & \multicolumn{1}{c|}{14.973} & \textbf{14.326}  \\
                         & $346\times 346$   & \multicolumn{1}{c|}{18.373} & \multicolumn{1}{c|}{18.374} & 18.374 & \multicolumn{1}{c|}{17.918} & \multicolumn{1}{c|}{17.617} & 17.016  & \multicolumn{1}{c|}{15.688} & \multicolumn{1}{c|}{\textbf{15.056}} & 14.313  \\
                         & $640\times 640$   & \multicolumn{1}{c|}{20.375} & \multicolumn{1}{c|}{20.376} & 20.375 & \multicolumn{1}{c|}{19.11} & \multicolumn{1}{c|}{18.47} & 17.464  & \multicolumn{1}{c|}{\textbf{15.719}} & \multicolumn{1}{c|}{14.997} & 14.221  \\
                         & $1280\times 1280$ & \multicolumn{1}{c|}{\textbf{24.093}} & \multicolumn{1}{c|}{\textbf{24.096}} & \textbf{24.101} & \multicolumn{1}{c|}{\textbf{20.362}} & \multicolumn{1}{c|}{\textbf{19.184}} & \textbf{17.731}  & \multicolumn{1}{c|}{15.699} & \multicolumn{1}{c|}{14.949} & 14.171  \\ \hline
        
                         \multirow{5}{*}{E2VID\cite{Rebecq19pami}}                     & $128\times 128$ & \multicolumn{1}{c|}{11.532} & \multicolumn{1}{c|}{11.557} & 11.528  & \multicolumn{1}{c|}{11.541} & \multicolumn{1}{c|}{11.532} & 11.501  & \multicolumn{1}{c|}{11.387} & \multicolumn{1}{c|}{11.268} & \textbf{10.972}  \\
                    & $240\times 240$   & \multicolumn{1}{c|}{12.186} & \multicolumn{1}{c|}{12.190} & 12.167  & \multicolumn{1}{c|}{12.168} & \multicolumn{1}{c|}{12.129} & \textbf{12.027}  & \multicolumn{1}{c|}{\textbf{11.629}} & \multicolumn{1}{c|}{\textbf{11.301}} & 10.871  \\
                    & $346\times 346$   & \multicolumn{1}{c|}{12.254} & \multicolumn{1}{c|}{12.249} & 12.260  & \multicolumn{1}{c|}{12.211} & \multicolumn{1}{c|}{\textbf{12.139}} & 11.965  & \multicolumn{1}{c|}{11.359} & \multicolumn{1}{c|}{11.009} & 10.531  \\
                    & $640\times 640$   & \multicolumn{1}{c|}{12.317} & \multicolumn{1}{c|}{12.293} & 12.316  & \multicolumn{1}{c|}{12.029} & \multicolumn{1}{c|}{11.754} & 11.396  & \multicolumn{1}{c|}{10.605} & \multicolumn{1}{c|}{10.254} & 9.858  \\
                    & $1280\times 1280$   & \multicolumn{1}{c|}{\textbf{13.379}} & \multicolumn{1}{c|}{\textbf{13.378}} & \textbf{13.381}  & \multicolumn{1}{c|}{\textbf{12.331}} & \multicolumn{1}{c|}{11.631} & 10.946  & \multicolumn{1}{c|}{10.050} & \multicolumn{1}{c|}{9.761} & 9.457  \\ \hline
                    \multirow{5}{*}{HF\cite{Scheerlinck18accv}}
                    & $128\times 128$ & \multicolumn{1}{c|}{11.143} & \multicolumn{1}{c|}{11.156} & 11.152  & \multicolumn{1}{c|}{11.146} & \multicolumn{1}{c|}{11.136} & 11.094  & \multicolumn{1}{c|}{10.972} & \multicolumn{1}{c|}{10.810} & \textbf{10.518}  \\
                    & $240\times 240$   & \multicolumn{1}{c|}{11.570} & \multicolumn{1}{c|}{11.569} & 11.562  & \multicolumn{1}{c|}{11.576} & \multicolumn{1}{c|}{11.572} & \textbf{11.512}  & \multicolumn{1}{c|}{\textbf{11.142}} & \multicolumn{1}{c|}{\textbf{10.892}} & 10.488  \\
                    & $346\times 346$   & \multicolumn{1}{c|}{11.606} & \multicolumn{1}{c|}{11.609} & 11.605  & \multicolumn{1}{c|}{11.599} & \multicolumn{1}{c|}{11.562} & 11.428  & \multicolumn{1}{c|}{10.976} & \multicolumn{1}{c|}{10.683} & 10.212  \\
                    & $640\times 640$   & \multicolumn{1}{c|}{11.646} & \multicolumn{1}{c|}{11.649} & 11.649  & \multicolumn{1}{c|}{11.552} & \multicolumn{1}{c|}{11.420} & 11.174  & \multicolumn{1}{c|}{10.433} & \multicolumn{1}{c|}{10.085} & 9.612  \\
                    & $1280\times 1280$   & \multicolumn{1}{c|}{\textbf{12.626}} & \multicolumn{1}{c|}{\textbf{12.609}} & \textbf{12.609}  & \multicolumn{1}{c|}{\textbf{12.041}} & \multicolumn{1}{c|}{\textbf{11.606}} & 11.013  & \multicolumn{1}{c|}{9.932} & \multicolumn{1}{c|}{9.591} & 9.142  \\\hline

    \end{tabular}}\caption{Image reconstruction evaluation on the \emph{carpet} sequence, at varying camera speeds and for model-based method HF \cite{Scheerlinck18accv} and learning-based method E2VID \cite{Rebecq19pami}. 
    In daytime and nighttime settings, increasing camera speed leads to a signficant drop in performance, especially for high-resolution sensors. This is why, in these settings, lower resolutions like $346\times 346$, $240\times 240$ and $128\times 128$ have higher performance, while having lower event-rate.}\label{tab:app:image_recons_comp}
\end{table*}

\begin{table*}[]
    \centering
    \resizebox{1\linewidth}{!}{%
    \begin{tabular}{lc|ccc|ccc|ccc}
        \hline
        \textbf{Dataset} & \textbf{Setting}  & \multicolumn{3}{c|}{\textbf{ideal}}                      & \multicolumn{3}{c|}{\textbf{daytime}}                         & \multicolumn{3}{c}{\textbf{nighttime}}                           \\ 
        &$f_\text{cutoff}$ [Hz]&\multicolumn{3}{c|}{$\infty$}&\multicolumn{3}{c|}{200}&\multicolumn{3}{c}{50}\\ \hline
                   & Velocity          & 1.2                     & 1.6                     & 2.4         & 1.2                     & 1.6                     & 2.4 & 1.2                     & 1.6                     & 2.4                    \\ \hline
                        \multirow{5}{*}{room}
& $128\times 128$   & \multicolumn{1}{c|}{2.36} & \multicolumn{1}{c|}{2.31} & 2.62 & \multicolumn{1}{c|}{2.21} & \multicolumn{1}{c|}{3.12} & 4.35 & \multicolumn{1}{c|}{8.54} & \multicolumn{1}{c|}{8.62} & 9.26 \\
& $240\times 240$   & \multicolumn{1}{c|}{1.26} & \multicolumn{1}{c|}{1.25} & 1.37 & \multicolumn{1}{c|}{1.76} & \multicolumn{1}{c|}{\textbf{2.39}} & \textbf{3.97} & \multicolumn{1}{c|}{6.96} & \multicolumn{1}{c|}{7.29} & 8.03 \\
& $346\times 346$   & \multicolumn{1}{c|}{1.03} & \multicolumn{1}{c|}{1.03} & 1.04 & \multicolumn{1}{c|}{1.84} & \multicolumn{1}{c|}{2.72} & 4.09 & \multicolumn{1}{c|}{6.68} & \multicolumn{1}{c|}{7.16} & 8.56 \\
& $640\times 640$   & \multicolumn{1}{c|}{\textbf{0.96}} & \multicolumn{1}{c|}{\textbf{0.92}} & \textbf{0.93} & \multicolumn{1}{c|}{\textbf{1.70}} & \multicolumn{1}{c|}{2.40} & 4.21 & \multicolumn{1}{c|}{6.32} & \multicolumn{1}{c|}{\textbf{6.75}} & 8.28 \\
& $1280\times 1280$ & \multicolumn{1}{c|}{1.33} & \multicolumn{1}{c|}{1.34} & 1.37 & \multicolumn{1}{c|}{1.95} & \multicolumn{1}{c|}{2.68} & 3.98 & \multicolumn{1}{c|}{\textbf{6.23}} & \multicolumn{1}{c|}{6.86} & \textbf{7.50} \\ \hline
    
                  & Velocity          & 0.9&1.2&1.8& 0.9&1.2&1.8& 0.9&1.2&1.8\\ \hline
\multirow{5}{*}{carpet}
& $128\times 128$   & \multicolumn{1}{c|}{2.09} & \multicolumn{1}{c|}{1.89} &2.11 & \multicolumn{1}{c|}{2.11} & \multicolumn{1}{c|}{1.92} &2.19 & \multicolumn{1}{c|}{\textbf{3.24}} & \multicolumn{1}{c|}{\textbf{3.59}} &  5.16\\
& $240\times 240$   & \multicolumn{1}{c|}{0.91} & \multicolumn{1}{c|}{0.93} &0.93 & \multicolumn{1}{c|}{0.75} & \multicolumn{1}{c|}{\textbf{0.95}} &\textbf{1.72} & \multicolumn{1}{c|}{3.69} & \multicolumn{1}{c|}{4.49} &  5.52\\
& $346\times 346$   & \multicolumn{1}{c|}{0.58} & \multicolumn{1}{c|}{0.62} &0.61 & \multicolumn{1}{c|}{\textbf{0.65}} & \multicolumn{1}{c|}{1.01} &1.82 & \multicolumn{1}{c|}{3.60} & \multicolumn{1}{c|}{4.51} &  5.85\\
& $640\times 640$   & \multicolumn{1}{c|}{\textbf{0.36}} & \multicolumn{1}{c|}{\textbf{0.37}} &\textbf{0.36} & \multicolumn{1}{c|}{0.88} & \multicolumn{1}{c|}{1.26} &1.91 & \multicolumn{1}{c|}{3.51} & \multicolumn{1}{c|}{4.55} & 5.87\\
& $1280\times 1280$ & \multicolumn{1}{c|}{0.48} & \multicolumn{1}{c|}{0.45} &0.46 & \multicolumn{1}{c|}{0.94} & \multicolumn{1}{c|}{1.32} &2.06 & \multicolumn{1}{c|}{3.48} & \multicolumn{1}{c|}{4.29} & \textbf{4.78} \\ \hline

                                   & Velocity          & 0.7                    & 1.0                     & 1.4& 0.7                    & 1.0                     & 1.4& 0.7                    & 1.0                     & 1.4\\ \hline
\multirow{5}{*}{rocks}
& $128\times 128$   & \multicolumn{1}{c|}{1.90} & \multicolumn{1}{c|}{1.56} & 1.64& \multicolumn{1}{c|}{2.10} & \multicolumn{1}{c|}{2.36} & 2.67 & \multicolumn{1}{c|}{\textbf{3.98}} & \multicolumn{1}{c|}{\textbf{4.96}} & \textbf{6.32}\\
& $240\times 240$   & \multicolumn{1}{c|}{0.74} & \multicolumn{1}{c|}{0.79} & 0.76& \multicolumn{1}{c|}{1.68} & \multicolumn{1}{c|}{\textbf{2.03}} & \textbf{2.60} & \multicolumn{1}{c|}{4.07} & \multicolumn{1}{c|}{5.35} & 6.39  \\
& $346\times 346$   & \multicolumn{1}{c|}{0.53} & \multicolumn{1}{c|}{0.47} & 0.50& \multicolumn{1}{c|}{1.74} & \multicolumn{1}{c|}{2.13} & 2.85 & \multicolumn{1}{c|}{4.24} & \multicolumn{1}{c|}{5.24} & 6.52  \\
& $640\times 640$   & \multicolumn{1}{c|}{\textbf{0.38}} & \multicolumn{1}{c|}{\textbf{0.37}} & \textbf{0.34}& \multicolumn{1}{c|}{1.47} & \multicolumn{1}{c|}{1.92} & 2.71 & \multicolumn{1}{c|}{4.81} & \multicolumn{1}{c|}{5.62} & 6.89  \\
& $1280\times 1280$ & \multicolumn{1}{c|}{0.42} & \multicolumn{1}{c|}{0.41} & 0.42& \multicolumn{1}{c|}{\textbf{1.43}} & \multicolumn{1}{c|}{1.93} & 3.04 & \multicolumn{1}{c|}{6.07} & \multicolumn{1}{c|}{6.78} & 7.59 \\ \hline

\end{tabular}}\caption{Optical flow accuracy of our Event-based Photometric Flow (EPF) following Eq. M-3.1. We report the resolution normalized end-point-error (RNEPE). 
All methods prefer higher resolution cameras in ideal and clean settings at almost all speeds. 
In noisy settings, EPF using high resolution event-cameras degrades, leading to lower resolution cameras being more accurate.}\label{tab:app:optical_flow}
\end{table*}

\begin{table*}[]
    \centering
    \resizebox{1\linewidth}{!}{%
    \begin{tabular}{lc|cccc|cccc|cccc}
        \hline
        \textbf{Method} & \textbf{Setting}  & \multicolumn{4}{c|}{\textbf{ideal}}                      & \multicolumn{4}{c|}{\textbf{daytime}}                         & \multicolumn{4}{c}{\textbf{nighttime}}                           \\ 
        &$f_\text{cutoff}$[Hz]&\multicolumn{4}{c|}{$\infty$}&\multicolumn{4}{c|}{$200$}&\multicolumn{4}{c}{50}\\\hline
          & Velocity  [m/s]        & 0.9                     & 1.2                     & 1.8         & 9& 0.9                     & 1.2                     & 1.8 & 9&0.9                     & 1.2                     & 1.8&9                    \\ \hline
\multirow{5}{*}{EPF (ours)}
                        & $128\times 128$   & \multicolumn{1}{c|}{2.09} & \multicolumn{1}{c|}{1.89} & \multicolumn{1}{c|}{2.11}  & \multicolumn{1}{c|}{1.87} & \multicolumn{1}{c|}{2.11} & \multicolumn{1}{c|}{1.92} & \multicolumn{1}{c|}{2.19} & \multicolumn{1}{c|}{5.09} & \multicolumn{1}{c|}{\textbf{3.24}} & \multicolumn{1}{c|}{\textbf{3.59}} & \multicolumn{1}{c|}{5.16} & \multicolumn{1}{c}{4.08} \\
                        & $240\times 240$   & \multicolumn{1}{c|}{0.91} & \multicolumn{1}{c|}{0.93} & \multicolumn{1}{c|}{0.93}  & \multicolumn{1}{c|}{0.89} & \multicolumn{1}{c|}{0.75} & \multicolumn{1}{c|}{\textbf{0.95}} & \multicolumn{1}{c|}{\textbf{1.72}} & \multicolumn{1}{c|}{5.84} & \multicolumn{1}{c|}{3.69} & \multicolumn{1}{c|}{4.49} & \multicolumn{1}{c|}{5.52} & \multicolumn{1}{c}{4.72} \\
                        & $346\times 346$   & \multicolumn{1}{c|}{0.58} & \multicolumn{1}{c|}{0.62} & \multicolumn{1}{c|}{0.61}  & \multicolumn{1}{c|}{0.62} & \multicolumn{1}{c|}{\textbf{0.65}} & \multicolumn{1}{c|}{1.01} & \multicolumn{1}{c|}{1.82} & \multicolumn{1}{c|}{6.10} & \multicolumn{1}{c|}{3.60} & \multicolumn{1}{c|}{4.51} & \multicolumn{1}{c|}{5.85} & \multicolumn{1}{c}{4.64} \\
                        & $640\times 640$   & \multicolumn{1}{c|}{\textbf{0.36}} & \multicolumn{1}{c|}{\textbf{0.37}} & \multicolumn{1}{c|}{\textbf{0.36}}  & \multicolumn{1}{c|}{\textbf{0.36}} & \multicolumn{1}{c|}{0.88} & \multicolumn{1}{c|}{1.26} & \multicolumn{1}{c|}{1.91} & \multicolumn{1}{c|}{6.19} & \multicolumn{1}{c|}{3.51} & \multicolumn{1}{c|}{4.55} & \multicolumn{1}{c|}{5.87} & \multicolumn{1}{c}{4.01} \\
                        & $1280\times 1280$ & \multicolumn{1}{c|}{0.48} & \multicolumn{1}{c|}{0.45} & \multicolumn{1}{c|}{0.46}  & \multicolumn{1}{c|}{0.48} & \multicolumn{1}{c|}{0.94} & \multicolumn{1}{c|}{1.32} & \multicolumn{1}{c|}{2.06} & \multicolumn{1}{c|}{\textbf{5.06}} & \multicolumn{1}{c|}{3.48} & \multicolumn{1}{c|}{4.29} & \multicolumn{1}{c|}{\textbf{4.78}} & \multicolumn{1}{c}{\textbf{3.75}} \\  \hline

                        \multirow{5}{*}{CM\cite{Gallego18cvpr}}
                        & $128\times 128$   & \multicolumn{1}{c|}{5.06} & \multicolumn{1}{c|}{4.94} & \multicolumn{1}{c|}{5.16}  & \multicolumn{1}{c|}{5.06} & \multicolumn{1}{c|}{4.89} & \multicolumn{1}{c|}{4.97}  & \multicolumn{1}{c|}{4.92} & \multicolumn{1}{c|}{4.92} & \multicolumn{1}{c|}{4.76} & \multicolumn{1}{c|}{4.78} & \multicolumn{1}{c|}{5.17} & \multicolumn{1}{c}{5.30} \\
                        & $240\times 240$   & \multicolumn{1}{c|}{2.51} & \multicolumn{1}{c|}{2.45} & \multicolumn{1}{c|}{2.49}  & \multicolumn{1}{c|}{2.51} &\multicolumn{1}{c|}{2.59} & \multicolumn{1}{c|}{2.61} & \multicolumn{1}{c|}{2.42}  & \multicolumn{1}{c|}{2.92} & \multicolumn{1}{c|}{ 2.57} & \multicolumn{1}{c|}{ 2.52} &  \multicolumn{1}{c|}{2.85} & \multicolumn{1}{c}{3.92} \\
                        & $346\times 346$   & \multicolumn{1}{c|}{1.74} & \multicolumn{1}{c|}{1.73} & \multicolumn{1}{c|}{1.78}  & \multicolumn{1}{c|}{1.72} &\multicolumn{1}{c|}{1.69} & \multicolumn{1}{c|}{1.75} & \multicolumn{1}{c|}{1.73}  & \multicolumn{1}{c|}{2.07} & \multicolumn{1}{c|}{ 1.72} & \multicolumn{1}{c|}{ 1.70} &  \multicolumn{1}{c|}{1.77} & \multicolumn{1}{c}{3.18} \\
                        & $640\times 640$   & \multicolumn{1}{c|}{0.71} & \multicolumn{1}{c|}{0.69} & \multicolumn{1}{c|}{0.72}  & \multicolumn{1}{c|}{0.69} &\multicolumn{1}{c|}{0.67} & \multicolumn{1}{c|}{0.66} & \multicolumn{1}{c|}{0.63}  & \multicolumn{1}{c|}{1.20} & \multicolumn{1}{c|}{ 0.69} & \multicolumn{1}{c|}{ 0.84} &  \multicolumn{1}{c|}{1.04} & \multicolumn{1}{c}{\textbf{3.09}} \\
                        & $1280\times 1280$ & \multicolumn{1}{c|}{\textbf{0.32}} & \multicolumn{1}{c|}{\textbf{0.32}} & \multicolumn{1}{c|}{\textbf{0.32}}  & \multicolumn{1}{c|}{\textbf{0.33}} &\multicolumn{1}{c|}{\textbf{0.25}} & \multicolumn{1}{c|}{\textbf{0.29}} & \multicolumn{1}{c|}{\textbf{0.30}}  & \multicolumn{1}{c|}{\textbf{0.88}} & \multicolumn{1}{c|}{\textbf{0.37}} & \multicolumn{1}{c|}{\textbf{0.45}} &  \multicolumn{1}{c|}{\textbf{0.63}} & \multicolumn{1}{c}{3.21} \\ \hline

                        \multirow{5}{*}{E-RAFT\cite{Gehrig21threedv}}
                        & $128\times 128$   & \multicolumn{1}{c|}{2.44} & \multicolumn{1}{c|}{2.87} & \multicolumn{1}{c|}{2.41} & \multicolumn{1}{c|}{2.91}                             & \multicolumn{1}{c|}{2.32} & \multicolumn{1}{c|}{3.04} & \multicolumn{1}{c|}{3.14}  & \multicolumn{1}{c|}{3.12} &\multicolumn{1}{c|}{2.56} & \multicolumn{1}{c|}{2.87} & \multicolumn{1}{c|}{2.98}&\multicolumn{1}{c}{7.97}   \\
                        & $240\times 240$   & \multicolumn{1}{c|}{1.01} & \multicolumn{1}{c|}{0.97} & \multicolumn{1}{c|}{1.01} & \multicolumn{1}{c|}{0.97} & \multicolumn{1}{c|}{1.10} & \multicolumn{1}{c|}{1.22} & \multicolumn{1}{c|}{1.37}  & \multicolumn{1}{c|}{2.04} &\multicolumn{1}{c|}{1.59} & \multicolumn{1}{c|}{1.71} & \multicolumn{1}{c|}{1.92}&\multicolumn{1}{c}{5.89}  \\
                        & $346\times 346$   & \multicolumn{1}{c|}{0.94} & \multicolumn{1}{c|}{0.85} & \multicolumn{1}{c|}{0.91} & \multicolumn{1}{c|}{0.92} & \multicolumn{1}{c|}{0.93} & \multicolumn{1}{c|}{0.99} & \multicolumn{1}{c|}{1.01}  & \multicolumn{1}{c|}{1.38} &\multicolumn{1}{c|}{1.24} & \multicolumn{1}{c|}{1.37} & \multicolumn{1}{c|}{1.47}&\multicolumn{1}{c}{4.68}  \\
                        & $640\times 640$   & \multicolumn{1}{c|}{0.79} & \multicolumn{1}{c|}{0.82} & \multicolumn{1}{c|}{0.76} & \multicolumn{1}{c|}{0.78} & \multicolumn{1}{c|}{0.84} & \multicolumn{1}{c|}{0.82} & \multicolumn{1}{c|}{0.85}  & \multicolumn{1}{c|}{1.31} &\multicolumn{1}{c|}{0.86} & \multicolumn{1}{c|}{1.02} & \multicolumn{1}{c|}{1.09}&\multicolumn{1}{c}{3.90}  \\
                        & $1280\times 1280$ & \multicolumn{1}{c|}{\textbf{0.61}} & \multicolumn{1}{c|}{\textbf{0.64}} & \multicolumn{1}{c|}{\textbf{0.63}} & \multicolumn{1}{c|}{\textbf{0.62}} & \multicolumn{1}{c|}{\textbf{0.63}} & \multicolumn{1}{c|}{\textbf{0.65}} & \multicolumn{1}{c|}{\textbf{0.65}}  & \multicolumn{1}{c|}{\textbf{1.03}} &\multicolumn{1}{c|}{\textbf{0.68}} & \multicolumn{1}{c|}{\textbf{0.75}} & \multicolumn{1}{c|}{\textbf{0.92}}&\multicolumn{1}{c}{\textbf{2.85}}  \\ \hline
    \end{tabular}}\caption{Optical flow evaluation on the \emph{carpet} sequence, at varying camera speeds, and for model-based method CM \cite{Gallego18cvpr} and learning-based method E-RAFT\cite{Gehrig21threedv}. 
    In daytime and nighttime settings, higher speeds significantly reduce performance for EPF, especially for high resolution cameras, making low resolution cameras a better choice.
    Nonetheless, both \cite{Gallego18cvpr} and \cite{Gehrig21threedv} remain robust, but with diminishing returns at higher resolutions.}\label{tab:app:optical_flow_sota}
\end{table*}

\begin{table*}[]
    \centering
    \begin{tabular}{lc|ccc|ccc|ccc}
        \hline
        \textbf{Dataset} & \textbf{Setting}  & \multicolumn{3}{c|}{\textbf{ideal}}                      & \multicolumn{3}{c|}{\textbf{daytime}}                         & \multicolumn{3}{c}{\textbf{nighttime}}                           \\ 
        &$f_\text{cutoff}$[Hz]&\multicolumn{3}{c|}{$\infty$}&\multicolumn{3}{c|}{200}&\multicolumn{3}{c}{50}\\\hline
        carpet           & Velocity          & 1.2                     & 1.6                     & 2.4         & 1.2                     & 1.6                     & 2.4 & 1.2                     & 1.6                     & 2.4                    \\ \hline
                            & $128\times 128$   & \multicolumn{1}{c|}{37.5}           & \multicolumn{1}{c|}{17.8}           & 32.7            & \multicolumn{1}{c|}{37.3}          & \multicolumn{1}{c|}{10.9}          & 4.88           & \multicolumn{1}{c|}{29.5}          & \multicolumn{1}{c|}{\textbf{2.44}} & 193.6  \\
                            & $240\times 240$   & \multicolumn{1}{c|}{17.1}           & \multicolumn{1}{c|}{18.2}           & 30.4            & \multicolumn{1}{c|}{21.6}          & \multicolumn{1}{c|}{20.4}          & 59.1           & \multicolumn{1}{c|}{33.0}          & \multicolumn{1}{c|}{59.6}          & \textbf{4.27}  \\
                            & $346\times 346$   & \multicolumn{1}{c|}{2.34}           & \multicolumn{1}{c|}{2.02}           & 2.43            & \multicolumn{1}{c|}{5.82}          & \multicolumn{1}{c|}{6.26}          & 11.1           & \multicolumn{1}{c|}{30.9}          & \multicolumn{1}{c|}{41.7}          & 11.6  \\
                            & $640\times 640$   & \multicolumn{1}{c|}{0.888}          & \multicolumn{1}{c|}{0.854}          & 0.959           & \multicolumn{1}{c|}{2.53}          & \multicolumn{1}{c|}{3.12}          & 6.84           & \multicolumn{1}{c|}{11.0}          & \multicolumn{1}{c|}{15.4}          & 102.0  \\
                            & $1280\times 1280$ & \multicolumn{1}{c|}{\textbf{0.617}} & \multicolumn{1}{c|}{\textbf{0.775}} & \textbf{0.661}  & \multicolumn{1}{c|}{\textbf{2.21}} & \multicolumn{1}{c|}{\textbf{2.81}} & \textbf{4.22}  & \multicolumn{1}{c|}{\textbf{6.39}} & \multicolumn{1}{c|}{11.5} & 24.9  \\ \hline
        rocks           & Velocity          & 1.2                     & 1.6                     & 2.4         & 1.2                     & 1.6                     & 2.4 & 1.2                     & 1.6                     & 2.4                    \\ \hline
                            & $128\times 128$   & \multicolumn{1}{c|}{14.0} & \multicolumn{1}{c|}{13.3} & 15.8  & \multicolumn{1}{c|}{18.5} & \multicolumn{1}{c|}{24.6} & 27.5  & \multicolumn{1}{c|}{38.4} & \multicolumn{1}{c|}{46.6} & 22.4  \\
                            & $240\times 240$   & \multicolumn{1}{c|}{3.84} & \multicolumn{1}{c|}{3.38} & 5.15  & \multicolumn{1}{c|}{7.17} & \multicolumn{1}{c|}{8.68} & 12.2  & \multicolumn{1}{c|}{17.2} & \multicolumn{1}{c|}{19.2} & 20.6  \\
                            & $346\times 346$   & \multicolumn{1}{c|}{1.63} & \multicolumn{1}{c|}{1.4} & 1.83  & \multicolumn{1}{c|}{5.81} & \multicolumn{1}{c|}{7.79} & 8.59  & \multicolumn{1}{c|}{\textbf{13.2}} & \multicolumn{1}{c|}{17.9} & 21.2  \\
                            & $640\times 640$   & \multicolumn{1}{c|}{1.18} & \multicolumn{1}{c|}{1.17} & 1.26  & \multicolumn{1}{c|}{5.3} & \multicolumn{1}{c|}{\textbf{7.54}} & \textbf{8.12}  & \multicolumn{1}{c|}{14.4} & \multicolumn{1}{c|}{\textbf{16.0}} & \textbf{18.2}  \\
                            & $1280\times 1280$ & \multicolumn{1}{c|}{\textbf{0.953}} & \multicolumn{1}{c|}{\textbf{0.925}} & \textbf{1.01}  & \multicolumn{1}{c|}{\textbf{5.0}} & \multicolumn{1}{c|}{8.88} & 9.42  & \multicolumn{1}{c|}{16.4} & \multicolumn{1}{c|}{19.1} & 31.8  \\ \hline
    \end{tabular}
    \caption{Effect of event camera resolution on median camera position accuracy (mm) with changing speed, and setting.}\label{tab:pose_tracking}
\end{table*}

\bibliographystyle{splncs04}
\bibliography{main}
\end{document}